\documentclass[11pt]{article}

\usepackage[preprint]{acl}
\usepackage{amssymb}
\usepackage{booktabs}
\usepackage[most]{tcolorbox}
\tcbuselibrary{listings} 
\usepackage{xcolor}   
\usepackage{array}
\usepackage{mathtools}

\newlength{\weightlhswidth}
\newcolumntype{L}[1]{>{\raggedright\arraybackslash}p{#1}}
\newcolumntype{C}[1]{>{\centering\arraybackslash}p{#1}}

\usepackage{times}
\usepackage{latexsym}

\usepackage{placeins}
\usepackage{makecell}
\usepackage{capt-of}
\usepackage{enumitem}

\usepackage{hyperref}

\usepackage{amsmath}
\allowdisplaybreaks

\usepackage[T1]{fontenc}

\usepackage[utf8]{inputenc}

\usepackage{microtype}

\usepackage{inconsolata}

\usepackage{graphicx}

\title{FACES: Fine-grained Gated Semantic Alignment for Multimodal Emotional Support Conversation}

\author{
  \textbf{Hang Ren, Peidong Wang, Yichen Gao, Wen Zhang}\\
  \textbf{Jiayang Yu, Xiaocui Yang, Shi Feng}$^{\dagger}$\\
  \normalfont School of Computer Science and Engineering,
  Northeastern University, China\\
  {\normalfont\large
    \texttt{\{renh2, gaoyc3, zhangw6\}@mails.neu.edu.cn}}\\
  {\normalfont\large
    \texttt{\{pdongwang, yujiayang\}@stumail.neu.edu.cn}}\\
  {\normalfont\large
    \texttt{\{yangxiaocui, fengshi\}@cse.neu.edu.cn}}
}

\begin{document}
\maketitle

\begingroup
  \renewcommand{\thefootnote}{\fnsymbol{footnote}}
  \footnotetext[2]{Corresponding author.}
\endgroup

\begin{abstract}
In recent years, with advances in multimodal interaction technologies and the growing demand for intelligent emotional support, Multimodal Emotional Support Conversation (MESC) has emerged as an important task. However, existing systems still suffer from shallow responses and inappropriate support strategies due to weak semantic feedback, cross-dimension reward compensation, and coarse reasoning supervision. To address these challenges, we propose \textsc{FACES}, a fine-grained gated semantic alignment framework for MESC. Specifically, \textsc{FACES} integrates multi-source rewards to construct multidimensional semantic and functional alignment signals, employs a non-compensatory gated fusion mechanism to reduce cross-dimension reward offset, and leverages structured Chain-of-Thought (CoT) anchors to perform token-level credit assignment. Under a controlled response-length protocol, FACES improves the average MLLM-based evaluation score from 3.61 to 4.04 and achieves an 80.22\% pairwise win rate over SOTA. Human evaluation under the same protocol yields a 75.3\% preference rate. Results on three external datasets further support transferability. The code and dataset are available at \url{https://github.com/OCT-ARR/arr}.
\end{abstract}

\section{Introduction}
\begin{figure}[t]
    \centering
    \includegraphics[width=\columnwidth]{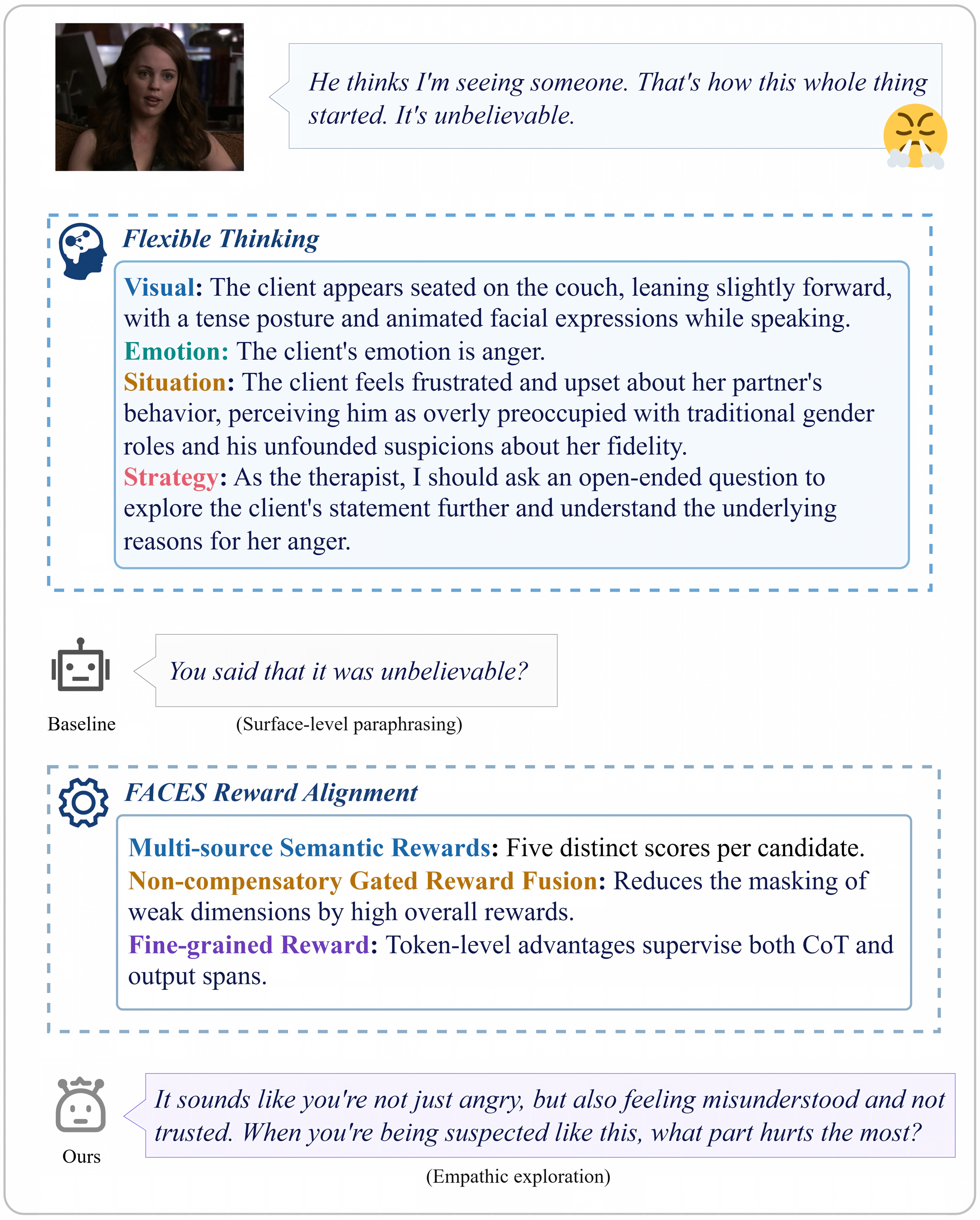}
    \caption{A MESC example. Both methods use flexible thinking to adaptively select contextually relevant reasoning aspects, while FACES adds reward alignment, recognizing the user's feeling of being misunderstood and asking a targeted follow-up question.}
    \label{fig:teaser_intro}
\end{figure}

Multimodal Emotional Support Conversation (MESC) \cite{chu2025towards} extends ESC \cite{liu2021towards} by generating empathetic, appropriate, and supportive responses from dialogue context and multimodal cues such as facial expressions, postures, and interaction context \cite{chaubey2026face,lu2025understanding}. Beyond producing a superficially natural response, MESC requires models to infer users' emotions, underlying causes, and support needs while grounding these in visual evidence. Correct emotion recognition alone does not ensure appropriate support \cite{liu2021towards,wang2024muffin}. In Figure~\ref{fig:teaser_intro}, the baseline identifies anger and relationship conflict but merely paraphrases the user, illustrating the gap between understanding and supportive action.

To better understand users and select support strategies, prior work has advanced MESC through dataset construction \cite{chu2025towards}, multimodal modeling \cite{le2026reinforce}, and reasoning-based training for support generation \cite{wang2025flexible}. Complementary work personalizes multimodal support through memory \cite{tu2025supportplay} and links emotion recognition and cause inference to support generation \cite{tu2026graph}. Reasoning-based methods use structured CoT \cite{wei2022chain} to make explicit how inferences about user emotions and underlying causes guide support strategy selection. In particular, FIRES \cite{wang2025flexible} introduces flexible thinking and shows that reinforcement learning (RL) objectives improve structured reasoning and final response quality in MESC.

Despite this progress, existing RL objectives for MESC still face three limitations in translating intermediate understanding into effective support. First, surface-overlap metrics such as BLEU and ROUGE-L \cite{papineni2002bleu,lin2004rouge} poorly capture cross-modal grounding, semantic consistency, and supportive intent, providing weak feedback on whether a response meets the user's support needs. Second, additive reward fusion can let strong understanding scores mask a poor support strategy or ineffective response \cite{van2013scalarized}. Accurate emotion or situation understanding alone cannot compensate for these support failures. Third, existing MESC methods relying on completion-level rewards \cite{wang2025flexible} lack stage-specific credit for visual grounding, emotion and situation understanding, strategy planning, or response generation. This coarse feedback makes it difficult to identify and correct the reasoning errors underlying an inappropriate support response.

To address these issues, we propose FACES, an MESC framework that uses gated multidimensional rewards to guide fine-grained token updates. Specifically, FACES uses CLIP \cite{radford2021learning} for visual perception, MiniLM \cite{wang2020minilm} for emotion understanding, and a multimodal large language model (MLLM) judge \cite{zheng2023judging,chen2024mllm} for situation understanding, support strategy, and response outcome. After a format-and-outcome warm-up, non-compensatory gated reward fusion limits compensation for weak strategy or response quality. Reasoning-stage markers route dimension-specific token-level advantages to the corresponding CoT and response spans, jointly optimizing intermediate reasoning and final response generation.

Our main contributions are as follows:
\begin{itemize}[leftmargin=1.1em,labelsep=0.4em,itemsep=0pt,topsep=1pt,parsep=0pt,partopsep=0pt]
    \item We propose a multi-source semantic reward scheme that provides more reliable semantic-level supervision than surface-overlap metrics.
    \item We introduce format-and-outcome warm-up with non-compensatory gated reward fusion to reduce cross-dimension compensation.
    \item We propose fine-grained semantic alignment rewards for structured CoT and outcome generation, aligning dimension-specific feedback with token spans to jointly optimize intermediate reasoning and the final response.
\end{itemize}

Under a 15-word response cap (cap-15) on MESC-FT, FACES raises the average MLLM-based evaluation score from 3.61 to 4.04, achieves an 80.22\% pairwise win rate, and obtains a 75.3\% human win rate over the SOTA baseline FIRES. Results on MELD-ESC300, StickerConv, and AvaMERG support cross-dataset transferability.

\section{Related Work}
\subsection{MESC Research}
Compared with traditional ESC \cite{liu2021towards}, MESC uses multimodal cues alongside dialogue, providing additional evidence about the help-seeker's emotion and situation when words are ambiguous \cite{chu2025towards,park2024av}. Two representative approaches have emerged. The first fuses multimodal signals to infer emotion and guide support generation, through sequential emotion--strategy--response modeling \cite{chu2025towards} or psychologically informed optimization \cite{le2026reinforce}. The second makes intermediate decisions explicit, using flexible reasoning over visual context, emotion, situation, and strategy \cite{wang2025flexible}, or appraisal-guided reasoning and retrieval \cite{chu2026beyond}. But remaining challenges include weak semantic feedback, compensation across reward dimensions, and insufficient supervision for intermediate reasoning; these are the problems FACES aims to address.

\subsection{Reward Evaluation and Fusion}
Reward evaluation and fusion aim to build reliable supervision and coordinate multiple objectives. Preference-based learning can optimize a policy through a learned reward model \cite{christiano2017deep,ouyang2022training,stiennon2020learning,bai2022constitutional}. Later work develops psychology-grounded, multi-perspective reward models \cite{wang2026perm}. MultiMood \cite{le2026reinforce} combines psychological criteria with PPO \cite{schulman2017proximal} and GRPO \cite{shao2024deepseekmath}; G-Eval \cite{liu2023g}, Prometheus \cite{kim2024prometheus}, PandaLM \cite{wang2024pandalm}, STICKERCONV \cite{zhang2024stickerconv}, and related studies \cite{zhao2024esc} demonstrate the utility of model-based evaluation. However, linear fusion can allow strong dimensions to mask weak ones \cite{van2013scalarized}, motivating non-compensatory reward fusion in multimodal emotional support conversations.

\subsection{Fine-grained Reward Optimization}

Fine-grained reward optimization moves beyond outcome-only feedback by supervising intermediate reasoning steps or local response segments. Prior work shows that process-level feedback improves reasoning reliability \cite{lightman2024lets}, and dense local signals are useful for multimodal alignment \cite{yu2024rlhf}. In ESC, FIRES optimizes flexible multimodal reasoning via completion-level rewards; RLVER derives emotion rewards from simulated users \citep{wang2025flexible,wang2026rlver}. Other methods provide finer-grained supervision at different levels: MindChat-R0 applies GRPO to structured empathetic reasoning; AFlow supervises dialogue prefixes; PEER trains a unified process--outcome reward model \citep{she2025mindchat,zou2026affective,wang2025peer}. In MESC, FACES uses non-compensatory fusion for heterogeneous dimension-level signals, routing their group-relative advantages to aligned token spans.

\section{Methodology}

\begin{figure*}[t]
    \centering
    \includegraphics[width=\textwidth]{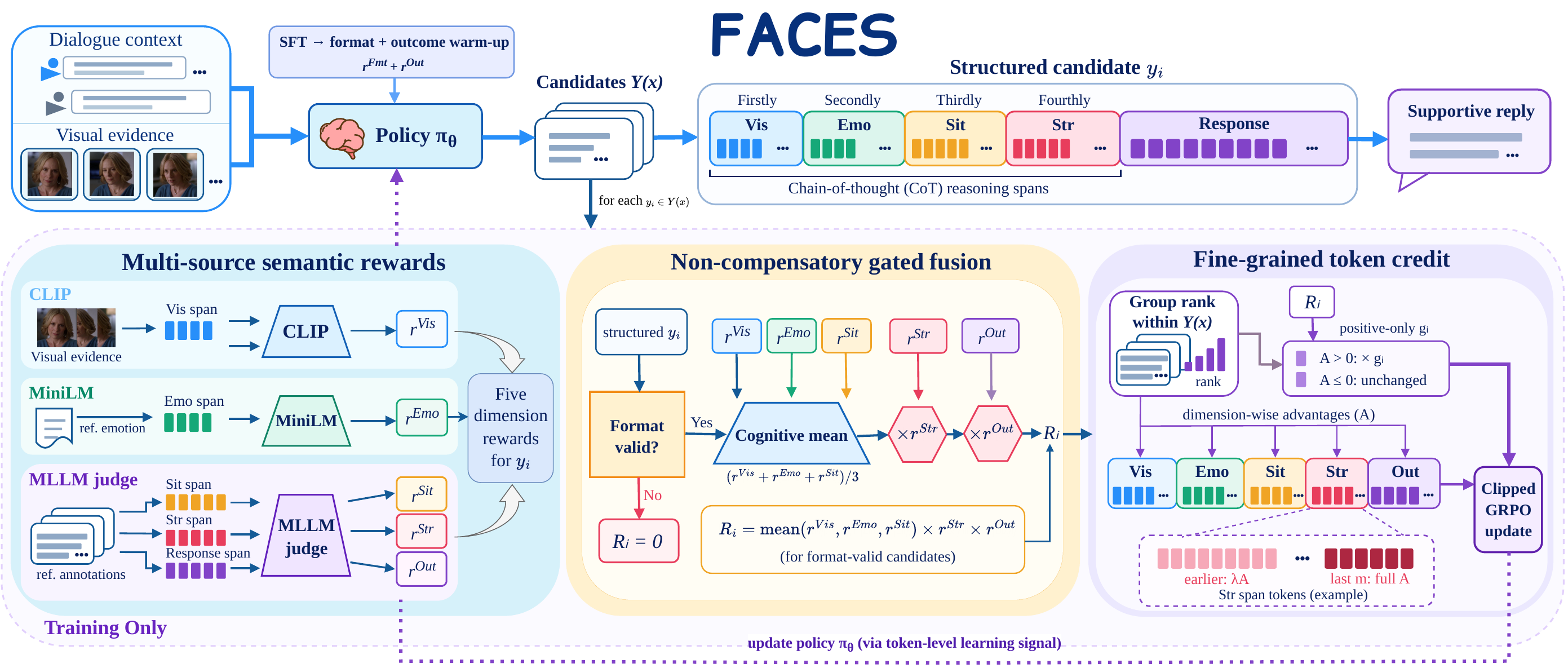}
    \caption{Workflow of FACES. After SFT and format and outcome warm-up, the policy samples structured CoT candidates \(y_i \in Y(x)\). CLIP and MiniLM score visual and emotion spans, while an MLLM judge (Qwen3-VL-8B-Instruct) scores situation, strategy, and outcome. Candidates failing the format check receive zero reward; valid candidates undergo non-compensatory fusion. Group ranking, positive-only advantage scaling, and token-level credit drive clipped GRPO updates for supportive responses.}
    \label{fig:work_flow}
\end{figure*}

To address the challenges outlined above, FACES optimizes structured candidates through the fine-grained reward workflow in Figure~\ref{fig:work_flow}, following supervised fine-tuning and a format-and-outcome RL warm-up. CLIP, MiniLM, and an MLLM judge provide five dimension-specific scores; non-compensatory fusion yields an overall reward that gates group-relative advantages, which are then assigned to aligned reasoning and response spans for targeted token-level updates to the policy.

\subsection{Multi-source Semantic Rewards}

For reward construction, BLEU and ROUGE-L primarily measure lexical overlap with reference responses \cite{papineni2002bleu,lin2004rouge} and may assign low scores to semantically valid empathetic replies with different wording.  To provide more reliable supervision, we construct multi-source rewards that extend beyond surface matching.

Given input $x$, the policy produces $\mathcal{Y}(x)=\{y_i\}_{i=1}^{N}$. We score five dimensions, $\mathcal{D}=\{\mathrm{Vis},\mathrm{Emo},\mathrm{Sit},\mathrm{Str},\mathrm{Out}\}$: visual perception, emotion understanding, situation understanding, support strategy, and response outcome. The first four dimensions follow the intermediate reasoning chain of visual perception, emotion recognition, situation attribution, and strategy planning. The outcome dimension separately captures final-response failures that may persist even when the intermediate reasoning appears plausible. This decomposition is guided by three criteria: it covers distinct MESC failure modes, aligns with explicit reasoning spans in the MESC-FT/FIRES flexible-thinking format \cite{wang2025flexible}, and supports both automatic scoring and model-based evaluation.

For visual perception and emotion understanding, we use CLIP and MiniLM. Let $z_i^{\mathrm{Vis}}$ and $z_i^{\mathrm{Emo}}$ denote the visual-perception and emotion-understanding spans extracted from $y_i$, let $I$ denote the input visual evidence, and let $z_*^{\mathrm{Emo}}$ denote the reference emotion span. The visual score is CLIP-based and binarized using a validation-selected threshold ($\tau_v=0.60$ by default; sensitivity is analyzed in Appendix~\ref{sec:appendix_sensitivity}):

\begin{equation}
r_i^{\mathrm{Vis}}=\mathbb{I}\!\left[\mathrm{CLIP}\!\left(z_i^{\mathrm{Vis}}, I\right)\ge \tau_v\right]
\label{eq:vis_score}
\end{equation}
The emotion score is computed with MiniLM as:
\begin{equation}
r_i^{\mathrm{Emo}}=\frac{1+\cos\!\left(e\!\left(z_i^{\mathrm{Emo}}\right),e\!\left(z_*^{\mathrm{Emo}}\right)\right)}{2}
\label{eq:emo_score}
\end{equation}
where $e(\cdot)$ denotes the MiniLM sentence encoder used for emotion representation; further implementation details are provided in Appendix~\ref{sec:appendix_reward_config}.

For the remaining semantic dimensions, $\mathcal{D}_{\mathrm{M}}=\{\mathrm{Sit},\mathrm{Str},\mathrm{Out}\}$, we use Qwen3-VL-8B-Instruct as the MLLM judge \cite{bai2025qwen3}. To reduce prompt sensitivity, we use three semantically equivalent prompts and score each candidate three times; a representative training prompt is provided in Appendix~\ref{sec:appendix_training_prompts}. The prompts jointly assess semantic consistency, functional consistency, and evidence-aware contextual fit, while penalizing overly verbose, repetitive, or formulaic outputs.

For each $d\in\mathcal{D}_{\mathrm{M}}$, the judge outputs a discrete score $s_{i,k}^{d}\in\{1,2,3,4,5\}$ for candidate $y_i$ under prompt variant $k$. We linearly rescale each score from the 1--5 range to $[0,1]$ via min--max normalization and then average across the three prompt variants to obtain the final score $r_i^{d}$. The reward system comprises a visual signal based on image--text alignment and four reference-conditioned text-based signals: Emotion uses the annotated emotion span, whereas Situation, Strategy, and Outcome use their corresponding reference annotations. Together, these signals form the basis for non-compensatory reward fusion and fine-grained credit assignment during policy optimization.

\subsection{Non-compensatory Gated Reward Fusion}

To reduce early optimization noise, we use a two-stage reward design. In warm-up, we optimize only format compliance and final response quality. Let $r_i^{\mathrm{Fmt}}\in\{0,1\}$ indicate whether $y_i$ satisfies the predefined structured format, and let $r_i^{\mathrm{Out}}\in[0,1]$ denote its response-quality score. The warm-up reward is defined as:

\begin{equation}
R_i^{\mathrm{Warm}} = r_i^{\mathrm{Fmt}} + r_i^{\mathrm{Out}}
\label{eq:warm_reward}
\end{equation}

This stage aims to build stable structural control and basic response quality before full RL.

Building on this initialization, we introduce non-compensatory gated fusion. Additive fusion provides a smooth signal but can let strong understanding scores mask poor strategy choices or ineffective responses, while multiplying all dimensions can make optimization overly sensitive to noise. We therefore separate cognitive dimensions, which offer complementary evidence about the user's state, from response-critical dimensions that directly affect support appropriateness and helpfulness.

Let the five dimension scores of candidate $y_i$ be $r_i^{\mathrm{Vis}}, r_i^{\mathrm{Emo}}, r_i^{\mathrm{Sit}}, r_i^{\mathrm{Str}}, r_i^{\mathrm{Out}}$, all normalized to $[0,1]$. This normalization aligns their numerical ranges but does not make their heterogeneous scales semantically equivalent; accordingly, we treat the fusion rule as an empirically motivated design choice. We first combine visual perception, emotion understanding, and situation understanding into a cognitive reward that reflects the inferred user state:

\begin{equation}
R_i^{\mathrm{Cog}}=
\frac{r_i^{\mathrm{Vis}} + r_i^{\mathrm{Emo}} + r_i^{\mathrm{Sit}}}{3}
\label{eq:cog_reward}
\end{equation}

We use mean fusion for the cognitive dimensions to reduce sensitivity to local uncertainty and avoid overly sparse supervision. In contrast, appropriate support strategy selection is central to effective emotional support \cite{liu2021towards,tu2022misc,cheng2022improving}, while response outcome directly reflects the final supportive quality of the reply. We thus use both as multiplicative gates.

The final reward is defined as:
\begin{equation}
R_i=
\begin{cases}
0, & r_i^{\mathrm{Fmt}} = 0,\\[4pt]
R_i^{\mathrm{Cog}} \cdot r_i^{\mathrm{Str}} \cdot r_i^{\mathrm{Out}}, & r_i^{\mathrm{Fmt}} = 1
\end{cases}
\label{eq:final_reward}
\end{equation}

The fusion averages the three cognitive scores and applies non-compensatory vetoes for zero Strategy or Outcome scores. For format-valid candidates, $R_i \le \min(R_i^{\mathrm{Cog}},r_i^{\mathrm{Str}},r_i^{\mathrm{Out}})$, so each factor bounds the total reward. Format-invalid candidates receive zero reward and zero token advantages.

\subsection{Fine-grained Rewards}

For each candidate response $y_i$, the preceding subsections yield dimension-wise scores $r_i^d$ and an overall reward $R_i$. Based on these signals, we convert candidate-level and segment-level supervision into token-level advantages aligned with the reasoning process, enabling joint optimization of intermediate reasoning and final responses.

As shown in Figure~\ref{fig:work_flow}, for each format-valid candidate $y_i$, we parse CoT and response spans $\mathcal{I}_i^d=[b_i^d,e_i^d)$, aligning the visual, emotion, situation, and strategy rewards with their respective CoT spans and $r_i^{\mathrm{Out}}$ with the final response.

Since absolute scores across dimensions may exhibit scale drift, we first convert them into centered rank-based advantages within each group. For any dimension $d \in \mathcal{D}$, we define:
\begin{equation}
\widetilde{A}_i^d=\frac{N+1}{2}-\operatorname{rank}(r_i^d),
\end{equation}
where $\operatorname{rank}(\cdot)$ denotes the descending rank within the group, with ties assigned the average rank.

To better align local updates with overall reward quality, we apply a simple positive gate based on the total reward. Let $R_{\max}$ denote the maximum total reward within the current group, and let $\gamma$ denote a validation-selected floor value (default 0.10; sensitivity analyzed in Appendix~\ref{sec:appendix_sensitivity}). The gate coefficient for candidate $i$ is defined as:
\begin{equation}
g_i=
\begin{cases}
\gamma, & R_{\max}\le \epsilon,\\[4pt]
\gamma+(1-\gamma)\dfrac{R_i}{R_{\max}}, & \text{otherwise},
\end{cases}
\end{equation}
where $\epsilon$ is a small constant. We then apply this gate only to positive segment advantages while leaving negative ones unchanged:
\begin{equation}
A_i^d=
\begin{cases}
g_i\,\widetilde{A}_i^d, & \widetilde{A}_i^d>0,\\[4pt]
\widetilde{A}_i^d, & \widetilde{A}_i^d\le 0.
\end{cases}
\end{equation}
This gate attenuates positive advantages from favorable local segments when a candidate's total reward is low relative to others in the same group.

For token-level credit assignment, we use an endpoint-anchored scheme. Segment endings often contain the semantic conclusion, while earlier tokens within the same span may encode the evidence, emotion judgment, and situation reasoning needed to reach that conclusion. We therefore avoid continuously decaying the weight with distance from the endpoint. Instead, we scale all earlier tokens by the same factor while retaining the full advantage for the final anchor tokens. Let $m$ be the anchor length and $\lambda$ the decay factor (defaults: $m=5$, $\lambda=0.8$; sensitivity is analyzed in Appendix~\ref{sec:appendix_sensitivity}). For any token $t\in\mathcal{I}_i^d$, we set:
\begin{equation}
A_{i,t}=
\begin{cases}
A_i^d, & e_i^d-m\le t<e_i^d,\\[4pt]
\lambda A_i^d, & b_i^d\le t<e_i^d-m.
\end{cases}
\end{equation}
Thus, the last $m$ tokens receive the full advantage, while earlier tokens receive a decayed one. For format-invalid candidates, whose segment boundaries cannot be reliably parsed, we simply set all token-level advantages to zero and do not apply reward-driven fine-grained supervision.

For policy optimization, FACES builds on the standard clipped GRPO update~\cite{shao2024deepseekmath}, replacing the completion-level advantage with our fine-grained token advantage $A_{i,t}$. Let $s_{i,t}=(x,y_{i,<t})$ and $T_i=|y_i|$. We denote by $\mathcal{S}_{i,t}^{\mathrm{clip}}(\theta)$ the clipped GRPO surrogate computed using the token-level probability ratio between the current policy and the rollout policy for $y_{i,t}$ at $s_{i,t}$, $\epsilon_{\mathrm{clip}}$, and $A_{i,t}$. The objective is:

\begin{equation}
\mathcal{L}_{\mathrm{fg}}(\theta)
=
-\frac{1}{N}\sum_{i=1}^{N}\frac{1}{T_i}
\sum_{t=1}^{T_i}
\left(
\mathcal{S}_{i,t}^{\mathrm{clip}}(\theta)
-
\beta \Delta_{i,t}^{\mathrm{KL}}
\right),
\label{eq:fg_grpo}
\end{equation}
where $\beta$ is the KL regularization coefficient, and $\Delta_{i,t}^{\mathrm{KL}}$ denotes the token-level KL penalty between the current policy $\pi_\theta$ and the reference policy $\pi_{\mathrm{ref}}$. In practice, $\pi_{\theta_{\mathrm{old}}}$ is kept fixed within the rollout batch, while $\pi_{\mathrm{ref}}$ is used only for KL regularization.

Overall, FACES combines group-relative advantages, total-reward-based gating, span-aligned token assignment, and clipped GRPO to optimize reasoning and response at the token level.

\begin{table*}[t]
\centering
\small
\setlength{\tabcolsep}{9pt}
\begin{tabular}{lcccccc}
\toprule
\textbf{Model} & \textbf{B-2 $\uparrow$} & \textbf{R-L $\uparrow$} & \textbf{M. $\uparrow$} & \textbf{F$_{\text{BERT}}$ $\uparrow$} & \textbf{Dist-3 $\uparrow$} & \textbf{DIFF-E. $\downarrow$} \\
\midrule
BlenderBot-Joint$^\ast$      & 4.85 & 15.25 & --    & 85.50 & --    & -- \\
BBMHR$^\ast$                 & 1.31 & 15.38 & --    & 86.60 & --    & -- \\

SMES$^\ast$                  & 5.13 & 15.42 & --    & 86.80 & --    & -- \\
Qwen2.5-VL$^\ast$            & 9.77 & 20.30 & 14.05 & 86.35 & 84.01 & 0.8744 \\
SMES\_Qwen2.5$^\ast$         & 9.24 & 19.88 & 13.16 & 85.62 & 92.61 & 0.7577 \\
FIRES (Baseline)$^\ast$      & \textbf{10.18} & \textbf{22.76} & 19.54 & 86.26 & 98.93 & 0.7209 \\
\midrule
GPT-4o                       & 2.84 & 9.97  & 16.56 & 95.75 & 59.36 & 0.5300 \\
Claude-Haiku-4.5            & 3.45 & 10.01 & 17.67 & 95.73 & 70.31 & 0.4753 \\
Gemini-2.5-Flash-Lite       & 4.81 & 12.86 & 15.90 & \textbf{96.41} & 67.69 & 0.4152 \\
GLM-4.1V-Thinking-FlashX    & 3.03 & 9.56  & 13.47 & 96.09 & 78.38 & 0.5316 \\
Doubao-1.6-Vision           & 2.86 & 10.81 & 14.86 & 96.01 & 77.08 & 0.4747 \\
Kimi-K2.5                   & 6.00 & 14.10 & 20.10 & 96.33 & 76.59 & 0.4417 \\
\midrule
GRPO (Format + Outcome)     & 9.93 & 21.48 & 18.05 & 86.33 & 95.17 & 0.4406 \\
Ours (Full)                 & 9.68 & 21.09 & \textbf{20.62} & 87.14 & \textbf{99.63} & \textbf{0.3643} \\
\bottomrule
\end{tabular}
\caption{Automatic evaluation results on the MESC-FT test set. The superscript $^\ast$ denotes results from prior work or corresponding baseline reports. ``--'' denotes unavailable results. Best results are shown in bold.}
\label{tab:automatic_evaluation}
\end{table*}

\section{Experiments}

\subsection{Experimental Setup}

{
\setlength{\parindent}{0.8em}

\noindent\textbf{Dataset.}
We conduct training and in-domain evaluation on MESC-FT, a FIRES-built dataset \cite{wang2025flexible} on top of MESC \cite{chu2025towards} for multimodal emotional support conversations, which provides instance-level flexible-thinking supervision. We use an approximately 8:1:1 split for training, validation, and testing. For cross-dataset evaluation, we construct MELD-ESC300, a 300-sample external test set derived from MELD \cite{poria2019meld}, and evaluate zero-shot transfer on the unseen multimodal benchmarks StickerConv \cite{zhang2024stickerconv} and AvaMERG \cite{zhang2025towards}. These datasets are used only for evaluation, without fine-tuning. Details of MELD-ESC300 construction and the StickerConv/AvaMERG evaluation protocols are provided in Appendix~\ref{sec:appendix_meld_esc300}.

\noindent\textbf{Evaluation Metrics.}
We evaluate responses using automatic metrics, MLLM judgments, and human preferences. Automatic metrics include BLEU-2 (B-2) \cite{papineni2002bleu}, ROUGE-L (R-L) \cite{lin2004rouge}, METEOR (M.) \cite{banerjee2005meteor}, BERTScore-F1 ($F_{\text{BERT}}$) \cite{zhang2020bertscore}, Distinct-3 (Dist-3) \cite{li2016diversity}, and DIFF-EPITOME (DIFF-E.) \cite{lee2022does}; higher is better except for DIFF-E. Following StickerConv's MLLM evaluation protocol \cite{zhang2024stickerconv}, we use Qwen2.5-VL-32B-Instruct-AWQ as the primary post-hoc judge \cite{bai2025qwen25vltechnicalreport}, distinct from the training-time reward judge. As a cross-family robustness check, GLM-4.6V-Flash judges the same fixed candidate pairs given identical dialogue context and visual evidence. For MESC-FT and MELD-ESC300, we report pointwise and blind pairwise MLLM evaluations with human pairwise preferences. For StickerConv and AvaMERG, we report automatic metrics and blind MLLM pairwise preferences without human annotation; protocols appear in Appendix~\ref{sec:appendix_external_generalization}.

\noindent\textbf{Compared Methods.}
We benchmark \textsc{FACES} against prior ESC/MESC systems \cite{liu2021towards,zhang2023ask,chu2025towards} and FIRES \cite{wang2025flexible}, as well as six strong, directly prompted MLLMs listed in Table~\ref{tab:automatic_evaluation}. All six receive the same dialogue context and visual evidence. To assess the three main components, we conduct contribution-level ablations and train DPO \cite{rafailov2023direct}, SimPO \cite{meng2024simpo}, and RLOO \cite{ahmadian2024back} as same-backbone baselines under identical SFT initialization and training prompts. Appendix~\ref{sec:appendix_additional_ablation} reports dimension-level reward ablations and offline diagnostics on gated reward fusion versus linear, geometric-mean, and Tchebycheff aggregation; Appendix~\ref{sec:appendix_sensitivity} tests endpoint-anchored token credit against linear and exponential decay; and Appendix~\ref{sec:appendix_setup} details baseline sources and implementation settings for reproducibility.
}

\begin{table*}[t]
\centering
\small
\setlength{\tabcolsep}{8pt}

{\footnotesize \textbf{(a)} Pointwise MLLM evaluation results on the MESC-FT test set. Scores are on a 1--5 scale, and higher is better.}\\[0.4em]
\begin{tabular}{lcccc}
\toprule
\textbf{Model} & \textbf{Empathy $\uparrow$} & \textbf{Multimodal App. $\uparrow$} & \textbf{Context-Response Cons. $\uparrow$} & \textbf{Avg. $\uparrow$} \\
\midrule
FIRES (Baseline) & 3.06 & 3.88 & 3.89 & 3.61 \\
Ours (Full)      & \textbf{3.98} & \textbf{4.03} & \textbf{4.11} & \textbf{4.04} \\
\bottomrule
\end{tabular}

\vspace{0.8em}

{\footnotesize \textbf{(b)} Pairwise MLLM preference evaluation results on the MESC-FT test set. Win rates are computed over 450 test instances.}\\[0.4em]
\begin{tabular}{llcccccc}
\toprule
\textbf{Model A} & \textbf{Model B} & \textbf{A Win} & \textbf{B Win} & \textbf{Tie} & \textbf{A Win Rate} & \textbf{B Win Rate} & \textbf{Tie Rate} \\
\midrule
Ours (Full) & FIRES (Baseline) & \textbf{361} & 53 & 36 & \textbf{80.22\%} & 11.78\% & 8.00\% \\
\bottomrule
\end{tabular}

\caption{Length-controlled MLLM evaluation results on the MESC-FT test set. Panel (a) reports pointwise scores across three evaluation dimensions, and panel (b) reports blind pairwise preferences over 450 test instances. Pairwise results are significant under a two-sided sign test with ties excluded ($p<0.001$).}
\label{tab:mllm_evaluation}
\end{table*}

\subsection{Main Results}
\subsubsection{Automatic Evaluation}

Table~\ref{tab:automatic_evaluation} shows that \textsc{FACES} leads on METEOR, Distinct-3, and DIFF-EPITOME, while FIRES leads on BLEU-2 and ROUGE-L and Gemini-2.5-Flash-Lite on BERTScore-F1. Although \textsc{FACES} trails FIRES on BLEU-2 and ROUGE-L, these metrics emphasize lexical overlap, and a supportive response need not closely match the reference wording. In contrast, its gains over FIRES on METEOR, BERTScore-F1, Distinct-3, and DIFF-EPITOME suggest stronger semantic correspondence, greater response diversity, and better support-strategy alignment. This pattern is consistent with semantic rewards favoring meaning over wording, while strategy and outcome gates discourage plausible but unhelpful replies.

\subsubsection{MLLM Evaluation}

To complement automatic metrics, we assess supportive quality with the MLLM judge using dialogue context and visual evidence. For brevity and symmetry, we remove reasoning traces and prefix-truncate \textsc{FACES} and baseline responses to 15 words; greedy decoding reduces sampling randomness. Details are in Appendix~\ref{sec:appendix_controlled_views}.

Under length control, the judge scores Empathy, Multimodal Appropriateness, and Context-Response Consistency on a 1--5 Likert scale, measuring overall emotional support quality, multimodal grounding, and contextual faithfulness, respectively. We average three judge prompts per metric to reduce prompt bias; detailed evaluation rules and prompts are in Appendix~\ref{sec:appendix_eval_prompts}.

Table~\ref{tab:mllm_evaluation}(a) shows that Ours (Full) improves all three pointwise dimensions over FIRES (Baseline) under cap-15, raising the average from 3.61 to 4.04. The largest gain is in Empathy, suggesting more supportive responses even with reasoning traces removed and responses truncated; other gains suggest better visual grounding and contextual consistency.

We further conduct blind pairwise preference evaluation under the same length-controlled setting. For each sample, the judge compares two anonymized responses given the same dialogue and visual context, selecting the better overall support response or a tie. We use three prompt variants with bidirectional candidate swapping, and majority voting reduces prompt and position bias.

Table~\ref{tab:mllm_evaluation}(b) reports an 80.22\% cap-15 win rate for Ours (Full) over FIRES (Baseline). Together with the pointwise gains, this suggests that response length alone does not explain the preference. Appendix~\ref{sec:appendix_cross_judge} reports an 84.9\% win rate for Ours (Full) under an independent GLM judge, indicating that the preference is not specific to the Qwen judge. Comparisons with directly prompted GPT-4o and five other strong MLLMs appear in Table~\ref{tab:main_gpt4o_comparison} and Appendix~\ref{sec:appendix_strong_mllm_pairwise}, respectively.

\begin{table}[t]
\centering
\scriptsize
\setlength{\tabcolsep}{3pt}
\begin{tabular}{llccc}
\toprule
\textbf{Dataset} & \textbf{Judge} & \textbf{FACES Win} & \textbf{GPT-4o Win} & \textbf{Tie} \\
\midrule
MESC-FT & Qwen2.5-VL-32B & 72.0\% & 16.0\% & 12.0\% \\
MESC-FT & GLM-4.6V-Flash & 69.8\% & 17.8\% & 12.4\% \\
MESC-FT & Human & 76.0\% & 14.0\% & 10.0\% \\
\midrule
MELD-ESC300 & Qwen2.5-VL-32B & 66.0\% & 14.0\% & 20.0\% \\
MELD-ESC300 & GLM-4.6V-Flash & 63.0\% & 16.7\% & 20.3\% \\
MELD-ESC300 & Human & 72.0\% & 13.3\% & 14.7\% \\
\bottomrule
\end{tabular}
\caption{Cap-15 pairwise comparison between \textsc{FACES} and directly prompted GPT-4o.}
\label{tab:main_gpt4o_comparison}
\end{table}

\begin{table}[t]
\centering
\small
\setlength{\tabcolsep}{7pt}
\begin{tabular}{lcc}
\toprule
\textbf{Dataset} & \textbf{Exact Agreement} & \textbf{Cohen's $\kappa$} \\
\midrule
MESC-FT & 74.7\% & 0.316 \\
MELD-ESC300 & 71.7\% & 0.441 \\
\bottomrule
\end{tabular}
\caption{Sample-level agreement between final human preferences and Qwen judge majority preferences for the FACES and FIRES comparison under cap-15.}
\label{tab:main_human_mllm_agreement}
\end{table}

\begin{table*}[t]
\centering
\small
\renewcommand{\arraystretch}{1.12}
\setlength{\tabcolsep}{7pt}
\begin{tabular}{lcccccc}
\toprule
\textbf{Model} & \textbf{B-2} $\uparrow$ & \textbf{R-L} $\uparrow$ & \textbf{M.} $\uparrow$ & \textbf{F$_{\text{BERT}}$} $\uparrow$ & \textbf{Dist-3} $\uparrow$ & \textbf{DIFF-E.} $\downarrow$ \\
\midrule
w/o multi-source reward & \textbf{10.46} & \textbf{22.31} & 19.67 & 86.55 & 98.62 & 0.4189 \\
w/o non-comp. gating    & 9.82 & 20.84 & 20.02 & 86.79 & 99.18 & 0.4057 \\
w/o fine-grained credit & 8.56 & 18.37 & 18.59 & 86.02 & 98.95 & 0.4902 \\
Ours (Full)             & 9.68 & 21.09 & \textbf{20.62} & \textbf{87.14} & \textbf{99.63} & \textbf{0.3643} \\
\bottomrule
\end{tabular}
\caption{Contribution-level ablation results on the MESC-FT test set. All metrics are higher-is-better except DIFF-E.}
\label{tab:ablation_study}
\end{table*}

\begin{table*}[t]
\centering
\small
\setlength{\tabcolsep}{12pt}
\begin{tabular}{lcccccc}
\toprule
\textbf{Model} & \textbf{B-2} $\uparrow$ & \textbf{R-L} $\uparrow$ & \textbf{M.} $\uparrow$ & \textbf{F$_{\text{BERT}}$} $\uparrow$ & \textbf{Dist-3} $\uparrow$ & \textbf{DIFF-E.} $\downarrow$ \\
\midrule
FIRES (Baseline) & \textbf{6.32} & \textbf{17.39} & 11.43 & 85.39 & 90.72 & 0.8192 \\
Ours (Full) & 5.59 & 16.57 & \textbf{15.16}$^\dagger$ & \textbf{86.23}$^\dagger$ & \textbf{94.07}$^\dagger$ & \textbf{0.7664}$^\dagger$ \\
\bottomrule
\end{tabular}
\caption{Automatic evaluation results on the MELD-ESC300 test set. All metrics are higher-is-better except DIFF-E. $^\dagger$ indicates a statistically significant improvement over FIRES under paired bootstrap resampling ($p<0.05$).}
\label{tab:generalization_auto}
\end{table*}

\subsubsection{Human Preference Evaluation}

To validate the cap-15 model-based evaluation, three annotators compared anonymized FACES and FIRES responses on all 450 MESC-FT and 300 MELD-ESC300 instances, assessing empathy, multimodal appropriateness, strategy suitability, and helpfulness using the same context and visual evidence. Candidate order was randomized and ties were allowed; majority vote determined labels, with a fourth adjudicator resolving three-way disagreements. Fleiss' $\kappa$ was computed from pre-adjudication labels. Appendix~\ref{sec:appendix_preference_eval} details annotator qualifications, protocol, and preference counts.

On MESC-FT, the FACES, FIRES, and tie rates are 75.3\%, 14.0\%, and 10.7\%; on MELD-ESC300, they are 70.3\%, 14.7\%, and 15.0\%. Both FACES advantages are statistically significant under two-sided sign tests excluding ties ($p<0.001$). Fleiss' $\kappa$ values are 0.62 and 0.58, respectively.

Table~\ref{tab:main_gpt4o_comparison} shows that the same annotators also prefer FACES to directly prompted GPT-4o in 76.0\% of MESC-FT and 72.0\% of MELD-ESC300 cases. Table~\ref{tab:main_human_mllm_agreement} reports exact human--Qwen agreement rates of 74.7\% and 71.7\%, with Cohen's $\kappa$ values of 0.316 and 0.441, respectively. The two evaluations agree beyond chance but still differ on some individual cases, as detailed in Appendix~\ref{sec:appendix_human_mllm_agreement}.

\subsection{Ablation Study}

To analyze the three main designs, we ablate multi-source rewards, non-compensatory gating, and fine-grained credit assignment from \textsc{FACES} using the same test set and evaluation protocol.

Table~\ref{tab:ablation_study} shows that Ours (Full) leads on METEOR, BERTScore-F1, Distinct-3, and DIFF-EPITOME, reflecting stronger semantic quality, diversity, and support-strategy alignment. Distinct metric declines across ablations reveal the components' complementary roles. Omitting multi-source rewards boosts BLEU-2 and ROUGE-L but lowers semantic and support-oriented metrics, consistent with overlap-based rewards favoring lexical matching. The overall profile weakens without non-compensatory gating; additive fusion allows cross-dimension compensation. Fine-grained credit assignment matters: its removal causes broad degradation, supporting span-aligned token-level supervision. Appendices~\ref{sec:appendix_additional_ablation} and \ref{sec:appendix_sensitivity} detail dimension-level ablations and hyperparameter sensitivity analyses.

\subsection{Generalization Analysis}

We assess transfer without further training on MELD-ESC300, StickerConv, and AvaMERG. MELD-ESC300 contains 300 samples from the MELD test split using EMOTyDA resources \cite{poria2019meld,saha2020towards}; a manual audit found that 274 (91.3\%) met the overall validity criterion. We evaluate all 300, treating the set as a controlled external test rather than a dedicated ESC benchmark. Appendix~\ref{sec:appendix_meld_esc300} details its construction.

Table~\ref{tab:generalization_auto} shows that Ours (Full) trails FIRES (Baseline) on BLEU-2 and ROUGE-L but improves METEOR, BERTScore-F1, Distinct-3, and DIFF-EPITOME, with all four gains significant under paired bootstrap resampling ($p<0.05$). This pattern suggests stronger semantic matching, diversity, and strategy alignment despite lower lexical overlap. Under cap-15, the MLLM average rises from 3.35 to 3.73 and the Qwen pairwise win rate reaches 62.00\%. Human preference reaches 70.3\% vs 14.7\% for FIRES ($p<0.001$, Fleiss' $\kappa=0.58$). Appendices~\ref{sec:appendix_meld_extra_auto}, \ref{sec:appendix_meld_mllm_eval}, and \ref{sec:appendix_human_preference} provide details.

On 300 StickerConv and 300 AvaMERG samples, Ours (Full) wins 63.7\% and 79.3\% of Qwen comparisons with FIRES under cap-15; the corresponding GLM rates are 64.3\% and 65.7\%. Appendices~\ref{sec:appendix_external_generalization} and \ref{sec:appendix_cross_judge} report the full results. Together, these tests support transfer across the evaluated domains, while Appendix~\ref{sec:appendix_case_study} shows a failure caused by incorrect visual grounding.

\begin{table}[t]
\centering
\small
\setlength{\tabcolsep}{12pt}
\begin{tabular}{lc}
\toprule
\textbf{Method} & \textbf{MLLM Avg. $\uparrow$} \\
\midrule
FIRES (Baseline) & 3.61 \\
DPO & 3.62 \\
SimPO & 3.77 \\
RLOO & 3.83 \\
Ours (Full) & \textbf{4.04} \\
\bottomrule
\end{tabular}
\caption{Comparison of optimization methods on MESC-FT under the same backbone, SFT initialization, prompts, and length-controlled MLLM protocol.}
\label{tab:optimization_baselines}
\end{table}

\subsection{Optimization}
To assess whether alternative alignment objectives yield comparable gains, we compare \textsc{FACES} with DPO, SimPO, and RLOO under the same backbone, SFT initialization, prompts, and evaluation protocol. Table~\ref{tab:optimization_baselines} shows that all three improve over FIRES but remain below \textsc{FACES}.

\section{Conclusion}
We present \textsc{FACES}, a fine-grained gated semantic alignment framework for MESC. It addresses weak semantic supervision, compensatory reward fusion, and coarse credit assignment through multidimensional rewards, non-compensatory gating, and span-aligned token-level advantages. On MESC-FT, \textsc{FACES} improves several automatic metrics and is preferred to FIRES in matched cap-15 MLLM and human evaluations. This preference advantage persists under an independent GLM judge, while ablations support the contributions of the three main designs. Evaluations on three external datasets provide further evidence of transfer without additional training. These findings highlight fine-grained semantic alignment with non-compensatory gating as a promising direction for MESC generation.

\section*{Limitations}

Although FACES shows consistent gains on multimodal emotional support generation, this study still leaves room for further extensions in personalization and evaluation. It focuses on instance-level semantic alignment and therefore does not explicitly model user-specific preferences. Moreover, our experimental setup prioritizes reproducibility and controlled comparison by relying on automatic metrics, MLLM-based assessment, pairwise human preference evaluation, and keyframe-based visual inputs. While these design choices are sufficient to establish the effectiveness of FACES in the present setting, future work could further strengthen the framework through finer-grained human evaluation, more diverse annotators, and richer multimodal signals such as vocal prosody and physiological signals in realistic support settings.

\section*{Ethical Considerations}

This study uses the MESC-FT dataset and constructs the MELD-ESC300 external generalization test set from publicly available MELD/EMOTyDA resources. All data are used solely for research purposes, and we strictly comply with the licensing, citation, and redistribution requirements of the original datasets. For released code, datasets, or derived resources, we avoid redistributing raw audio, video, or image materials without proper authorization; instead, we provide guidance for obtaining the original datasets, along with data processing scripts and reproduction instructions.

This work focuses on multimodal emotional support conversation generation, but the proposed model remains a research prototype. It should not be used as a tool for clinical diagnosis, psychotherapy, or crisis intervention without professional supervision. The model may inherit biases from the training data and foundation models, which could lead to inappropriate judgments in emotion recognition, support strategy selection, or response generation. For acute psychological crises, personal safety risks, or other scenarios requiring professional intervention, any real-world deployment should incorporate human oversight, safety screening, and professional referral protocols to determine whether and how the system should respond.

\bibliography{custom}

\begin{thebibliography}{53}
\providecommand{\natexlab}[1]{#1}

\bibitem[{Ahmadian et~al.(2024)Ahmadian, Cremer, Gall{\'e}, Fadaee, Kreutzer, Pietquin, Ustun, and Hooker}]{ahmadian2024back}
Arash Ahmadian, Chris Cremer, Matthias Gall{\'e}, Marzieh Fadaee, Julia Kreutzer, Olivier Pietquin, Ahmet Ustun, and Sara Hooker. 2024.
\newblock Back to basics: Revisiting reinforce style optimization for learning from human feedback in llms.
\newblock In \emph{Proceedings of the 62nd Annual Meeting of the Association for Computational Linguistics}.

\bibitem[{Bai et~al.(2025{\natexlab{a}})Bai, Cai, Chen, Chen, Chen, Cheng, Deng, Ding, Gao, Ge, Ge, Guo, Huang, Huang, Huang, Hui, Jiang, Li, Li, Li, Li, Lin, Lin, Liu, Liu, Liu, Liu, Liu, Liu, Lu, Luo, Lv, Men, Meng, Ren, Ren, Song, Sun, Tang, Tu, Wan, Wang, Wang, Wang, Wang, Xie, Xu, Xu, Xu, Yang, Yang, Yang, Yang, Yu, Zhang, Zhang, Zhang, Zheng, Zhong, Zhou, Zhou, Zhou, Zhu, and Zhu}]{bai2025qwen3}
Shuai Bai, Yuxuan Cai, Ruizhe Chen, Keqin Chen, Xionghui Chen, Zesen Cheng, Lianghao Deng, Wei Ding, Chang Gao, Chunjiang Ge, Wenbin Ge, Zhifang Guo, Qidong Huang, Jie Huang, Fei Huang, Binyuan Hui, Shutong Jiang, Zhaohai Li, Mingsheng Li, and 45 others. 2025{\natexlab{a}}.
\newblock Qwen3-{VL} technical report.
\newblock \emph{arXiv preprint arXiv:2511.21631}.

\bibitem[{Bai et~al.(2025{\natexlab{b}})Bai, Chen, Liu, Wang, Ge, Song, Dang, Wang, Wang, Tang, Zhong, Zhu, Yang, Li, Wan, Wang, Ding, Fu, Xu, Ye, Zhang, Xie, Cheng, Zhang, Yang, Xu, and Lin}]{bai2025qwen25vltechnicalreport}
Shuai Bai, Keqin Chen, Xuejing Liu, Jialin Wang, Wenbin Ge, Sibo Song, Kai Dang, Peng Wang, Shijie Wang, Jun Tang, Humen Zhong, Yuanzhi Zhu, Mingkun Yang, Zhaohai Li, Jianqiang Wan, Pengfei Wang, Wei Ding, Zheren Fu, Yiheng Xu, and 8 others. 2025{\natexlab{b}}.
\newblock Qwen2.5-{VL} technical report.
\newblock \emph{arXiv preprint arXiv:2502.13923}.

\bibitem[{Bai et~al.(2022)Bai, Kadavath, Kundu, Askell, Kernion, Jones, Chen, Goldie, Mirhoseini, McKinnon, Chen, Olsson, Olah, Hernandez, Drain, Ganguli, Li, Tran-Johnson, Perez, Kerr, Mueller, Ladish, Landau, Ndousse, Lukosuite, Lovitt, Sellitto, Elhage, Schiefer, Mercado, DasSarma, Lasenby, Larson, Ringer, Johnston, Kravec, El~Showk, Fort, Lanham, Telleen-Lawton, Conerly, Henighan, Hume, Bowman, Hatfield-Dodds, Mann, Amodei, Joseph, McCandlish, Brown, and Kaplan}]{bai2022constitutional}
Yuntao Bai, Saurav Kadavath, Sandipan Kundu, Amanda Askell, Jackson Kernion, Andy Jones, Anna Chen, Anna Goldie, Azalia Mirhoseini, Cameron McKinnon, Carol Chen, Catherine Olsson, Christopher Olah, Danny Hernandez, Dawn Drain, Deep Ganguli, Dustin Li, Eli Tran-Johnson, Ethan Perez, and 32 others. 2022.
\newblock Constitutional {AI}: Harmlessness from {AI} feedback.
\newblock \emph{arXiv preprint arXiv:2212.08073}.

\bibitem[{Banerjee and Lavie(2005)}]{banerjee2005meteor}
Satanjeev Banerjee and Alon Lavie. 2005.
\newblock Meteor: An automatic metric for mt evaluation with improved correlation with human judgments.
\newblock In \emph{Proceedings of the acl workshop on intrinsic and extrinsic evaluation measures for machine translation and/or summarization}, pages 65--72.

\bibitem[{Chaubey et~al.(2026)Chaubey, Guan, and Soleymani}]{chaubey2026face}
Ashutosh Chaubey, Xulang Guan, and Mohammad Soleymani. 2026.
\newblock Face-llava: Facial expression and attribute understanding through instruction tuning.
\newblock In \emph{Proceedings of the IEEE/CVF Winter Conference on Applications of Computer Vision}, pages 2648--2660.

\bibitem[{Chen et~al.(2024)Chen, Chen, Zhang, Wang, Liu, Zhou, Zhang, Wan, Zhou, and Sun}]{chen2024mllm}
Dongping Chen, Ruoxi Chen, Shilin Zhang, Yaochen Wang, Yinuo Liu, Huichi Zhou, Qihui Zhang, Yao Wan, Pan Zhou, and Lichao Sun. 2024.
\newblock Mllm-as-a-judge: Assessing multimodal llm-as-a-judge with vision-language benchmark.
\newblock In \emph{Forty-first International Conference on Machine Learning}.

\bibitem[{Cheng et~al.(2022)Cheng, Liu, Li, Wang, Zhao, Liu, Liang, and Zheng}]{cheng2022improving}
Yi~Cheng, Wenge Liu, Wenjie Li, Jiashuo Wang, Ruihui Zhao, Bang Liu, Xiaodan Liang, and Yefeng Zheng. 2022.
\newblock Improving multi-turn emotional support dialogue generation with lookahead strategy planning.
\newblock In \emph{Proceedings of the 2022 Conference on Empirical Methods in Natural Language Processing}.

\bibitem[{Christiano et~al.(2017)Christiano, Leike, Brown, Martic, Legg, and Amodei}]{christiano2017deep}
Paul~F Christiano, Jan Leike, Tom Brown, Miljan Martic, Shane Legg, and Dario Amodei. 2017.
\newblock Deep reinforcement learning from human preferences.
\newblock \emph{Advances in neural information processing systems}, 30.

\bibitem[{Chu et~al.(2026)Chu, Liao, Liang, Li, and Hong}]{chu2026beyond}
Yuqi Chu, Lizi Liao, Jinggui Liang, Boyang Li, and Richang Hong. 2026.
\newblock Beyond semantic similarity: Appraisal-guided {Chain-of-Thought} reasoning and retrieval for multimodal emotional support conversations.
\newblock In \emph{Findings of the Association for Computational Linguistics: ACL 2026}.

\bibitem[{Chu et~al.(2025)Chu, Liao, Zhou, Ngo, and Hong}]{chu2025towards}
Yuqi Chu, Lizi Liao, Zhiyuan Zhou, Chong-Wah Ngo, and Richang Hong. 2025.
\newblock Towards multimodal emotional support conversation systems.
\newblock \emph{IEEE Transactions on Multimedia}.

\bibitem[{Kim et~al.(2024)Kim, Shin, Cho, Jang, Longpre, Lee, Yun, Shin, Kim, Thorne, and Seo}]{kim2024prometheus}
Seungone Kim, Jamin Shin, Yejin Cho, Joel Jang, Shayne Longpre, Hwaran Lee, Sangdoo Yun, Seongjin Shin, Sungdong Kim, James Thorne, and Minjoon Seo. 2024.
\newblock Prometheus: Inducing fine-grained evaluation capability in language models.
\newblock In \emph{The Twelfth International Conference on Learning Representations}.

\bibitem[{Le et~al.(2026)Le, Nguyen, Vo, Nguyen, Le~Binh, Nguyen, Sonntag, Liao, and Nguyen}]{le2026reinforce}
Huy~M Le, Dat~Tien Nguyen, Ngan~TT Vo, Tuan~DQ Nguyen, Nguyen Le~Binh, Duy Minh~Ho Nguyen, Daniel Sonntag, Lizi Liao, and Binh~T Nguyen. 2026.
\newblock Reinforce trustworthiness in multimodal emotional support system.
\newblock In \emph{Proceedings of the AAAI Conference on Artificial Intelligence}, volume~40, pages 31474--31482.

\bibitem[{Lee et~al.(2022)Lee, Lim, and Choi}]{lee2022does}
Young-Jun Lee, Chae-Gyun Lim, and Ho-Jin Choi. 2022.
\newblock Does gpt-3 generate empathetic dialogues? a novel in-context example selection method and automatic evaluation metric for empathetic dialogue generation.
\newblock In \emph{Proceedings of the 29th international conference on computational linguistics}, pages 669--683.

\bibitem[{Li et~al.(2016)Li, Galley, Brockett, Gao, and Dolan}]{li2016diversity}
Jiwei Li, Michel Galley, Chris Brockett, Jianfeng Gao, and William~B Dolan. 2016.
\newblock A diversity-promoting objective function for neural conversation models.
\newblock In \emph{Proceedings of the 2016 conference of the North American chapter of the association for computational linguistics: human language technologies}, pages 110--119.

\bibitem[{Lightman et~al.(2024)Lightman, Kosaraju, Burda, Edwards, Baker, Lee, Leike, Schulman, Sutskever, and Cobbe}]{lightman2024lets}
Hunter Lightman, Vineet Kosaraju, Yuri Burda, Harrison Edwards, Bowen Baker, Teddy Lee, Jan Leike, John Schulman, Ilya Sutskever, and Karl Cobbe. 2024.
\newblock Let's verify step by step.
\newblock In \emph{The Twelfth International Conference on Learning Representations}.

\bibitem[{Lin(2004)}]{lin2004rouge}
Chin-Yew Lin. 2004.
\newblock Rouge: A package for automatic evaluation of summaries.
\newblock In \emph{Text summarization branches out}, pages 74--81.

\bibitem[{Liu et~al.(2021)Liu, Zheng, Demasi, Sabour, Li, Yu, Jiang, and Huang}]{liu2021towards}
Siyang Liu, Chujie Zheng, Orianna Demasi, Sahand Sabour, Yu~Li, Zhou Yu, Yong Jiang, and Minlie Huang. 2021.
\newblock Towards emotional support dialog systems.
\newblock In \emph{Proceedings of the 59th annual meeting of the association for computational linguistics and the 11th international joint conference on natural language processing (volume 1: Long papers)}, pages 3469--3483.

\bibitem[{Liu et~al.(2023)Liu, Iter, Xu, Wang, Xu, and Zhu}]{liu2023g}
Yang Liu, Dan Iter, Yichong Xu, Shuohang Wang, Ruochen Xu, and Chenguang Zhu. 2023.
\newblock G-eval: Nlg evaluation using gpt-4 with better human alignment.
\newblock In \emph{Proceedings of the 2023 conference on empirical methods in natural language processing}, pages 2511--2522.

\bibitem[{Lu et~al.(2025)Lu, Chen, Liang, Tan, Zeng, and Hu}]{lu2025understanding}
Haifeng Lu, Jiuyi Chen, Feng Liang, Mingkui Tan, Runhao Zeng, and Xiping Hu. 2025.
\newblock Understanding emotional body expressions via large language models.
\newblock In \emph{Proceedings of the AAAI Conference on Artificial Intelligence}, volume~39, pages 1447--1455.

\bibitem[{Meng et~al.(2024)Meng, Xia, and Chen}]{meng2024simpo}
Yu~Meng, Mengzhou Xia, and Danqi Chen. 2024.
\newblock Simpo: Simple preference optimization with a reference-free reward.
\newblock In \emph{Advances in Neural Information Processing Systems}.

\bibitem[{Ouyang et~al.(2022)Ouyang, Wu, Jiang, Almeida, Wainwright, Mishkin, Zhang, Agarwal, Slama, Ray, Schulman, Hilton, Kelton, Miller, Simens, Askell, Welinder, Christiano, Leike, and Lowe}]{ouyang2022training}
Long Ouyang, Jeffrey Wu, Xu~Jiang, Diogo Almeida, Carroll~L. Wainwright, Pamela Mishkin, Chong Zhang, Sandhini Agarwal, Katarina Slama, Alex Ray, John Schulman, Jacob Hilton, Fraser Kelton, Luke Miller, Maddie Simens, Amanda Askell, Peter Welinder, Paul Christiano, Jan Leike, and Ryan Lowe. 2022.
\newblock Training language models to follow instructions with human feedback.
\newblock \emph{Advances in neural information processing systems}, 35:27730--27744.

\bibitem[{Papineni et~al.(2002)Papineni, Roukos, Ward, and Zhu}]{papineni2002bleu}
Kishore Papineni, Salim Roukos, Todd Ward, and Wei-Jing Zhu. 2002.
\newblock Bleu: a method for automatic evaluation of machine translation.
\newblock In \emph{Proceedings of the 40th annual meeting of the Association for Computational Linguistics}, pages 311--318.

\bibitem[{Park et~al.(2024)Park, Kim, Rha, Godiva, and Ro}]{park2024av}
Se~Jin Park, Yeonju Kim, Hyeongseop Rha, Bella Godiva, and Yong~Man Ro. 2024.
\newblock Av-emodialog: Chat with audio-visual users leveraging emotional cues.
\newblock \emph{arXiv preprint arXiv:2412.17292}.

\bibitem[{Poria et~al.(2019)Poria, Hazarika, Majumder, Naik, Cambria, and Mihalcea}]{poria2019meld}
Soujanya Poria, Devamanyu Hazarika, Navonil Majumder, Gautam Naik, Erik Cambria, and Rada Mihalcea. 2019.
\newblock Meld: A multimodal multi-party dataset for emotion recognition in conversations.
\newblock In \emph{Proceedings of the 57th annual meeting of the association for computational linguistics}, pages 527--536.

\bibitem[{Radford et~al.(2021)Radford, Kim, Hallacy, Ramesh, Goh, Agarwal, Sastry, Askell, Mishkin, Clark, Krueger, and Sutskever}]{radford2021learning}
Alec Radford, Jong~Wook Kim, Chris Hallacy, Aditya Ramesh, Gabriel Goh, Sandhini Agarwal, Girish Sastry, Amanda Askell, Pamela Mishkin, Jack Clark, Gretchen Krueger, and Ilya Sutskever. 2021.
\newblock Learning transferable visual models from natural language supervision.
\newblock In \emph{Proceedings of the 38th International Conference on Machine Learning}, volume 139 of \emph{Proceedings of Machine Learning Research}, pages 8748--8763. PMLR.

\bibitem[{Rafailov et~al.(2023)Rafailov, Sharma, Mitchell, Manning, Ermon, and Finn}]{rafailov2023direct}
Rafael Rafailov, Archit Sharma, Eric Mitchell, Christopher~D Manning, Stefano Ermon, and Chelsea Finn. 2023.
\newblock Direct preference optimization: Your language model is secretly a reward model.
\newblock \emph{Advances in neural information processing systems}, 36:53728--53741.

\bibitem[{Saha et~al.(2020)Saha, Patra, Saha, and Bhattacharyya}]{saha2020towards}
Tulika Saha, Aditya Patra, Sriparna Saha, and Pushpak Bhattacharyya. 2020.
\newblock Towards emotion-aided multi-modal dialogue act classification.
\newblock In \emph{Proceedings of the 58th annual meeting of the association for computational linguistics}, pages 4361--4372.

\bibitem[{Schulman et~al.(2017)Schulman, Wolski, Dhariwal, Radford, and Klimov}]{schulman2017proximal}
John Schulman, Filip Wolski, Prafulla Dhariwal, Alec Radford, and Oleg Klimov. 2017.
\newblock Proximal policy optimization algorithms.
\newblock \emph{arXiv preprint arXiv:1707.06347}.

\bibitem[{Shao et~al.(2024)Shao, Wang, Zhu, Xu, Song, Bi, Zhang, Zhang, Li, Wu, and Guo}]{shao2024deepseekmath}
Zhihong Shao, Peiyi Wang, Qihao Zhu, Runxin Xu, Junxiao Song, Xiao Bi, Haowei Zhang, Mingchuan Zhang, Y.~K. Li, Y.~Wu, and Daya Guo. 2024.
\newblock {DeepSeekMath}: Pushing the limits of mathematical reasoning in open language models.
\newblock \emph{arXiv preprint arXiv:2402.03300}.

\bibitem[{She et~al.(2025)She, Zhang, Yao, Gao, and Jin}]{she2025mindchat}
Dong She, Chenxu Zhang, Xianrong Yao, Yang Gao, and Zhanpeng Jin. 2025.
\newblock {MindChat-R0}: A large language model for emotionally supportive dialogue through reinforcement learning.
\newblock In \emph{Companion of the 2025 ACM International Joint Conference on Pervasive and Ubiquitous Computing}.

\bibitem[{Stiennon et~al.(2020)Stiennon, Ouyang, Wu, Ziegler, Lowe, Voss, Radford, Amodei, and Christiano}]{stiennon2020learning}
Nisan Stiennon, Long Ouyang, Jeffrey Wu, Daniel Ziegler, Ryan Lowe, Chelsea Voss, Alec Radford, Dario Amodei, and Paul~F Christiano. 2020.
\newblock Learning to summarize with human feedback.
\newblock \emph{Advances in neural information processing systems}, 33:3008--3021.

\bibitem[{Tu et~al.(2026)Tu, Niu, Zeng, Xu, and Zhang}]{tu2026graph}
Geng Tu, Taiyu Niu, Xi~Zeng, Ruifeng Xu, and Min Zhang. 2026.
\newblock A graph-enhanced {MLLM} for hierarchical multimodal emotion understanding and support in conversations.
\newblock In \emph{Proceedings of the 49th International ACM SIGIR Conference on Research and Development in Information Retrieval}.

\bibitem[{Tu et~al.(2025)Tu, Wang, Cambria, Li, and Xu}]{tu2025supportplay}
Geng Tu, Bingbing Wang, Erik Cambria, Wenjie Li, and Ruifeng Xu. 2025.
\newblock {SupportPlay}: A multi-agent role-playing system for personalized and sustained multimodal emotional support conversation.
\newblock In \emph{Companion Proceedings of the ACM on Web Conference 2025}.

\bibitem[{Tu et~al.(2022)Tu, Li, Cui, Wang, Wen, and Yan}]{tu2022misc}
Quan Tu, Yanran Li, Jianwei Cui, Bin Wang, Ji-Rong Wen, and Rui Yan. 2022.
\newblock Misc: A mixed strategy-aware model integrating comet for emotional support conversation.
\newblock In \emph{Proceedings of the 60th annual meeting of the association for computational linguistics (volume 1: Long papers)}, pages 308--319.

\bibitem[{Van~Moffaert et~al.(2013)Van~Moffaert, Drugan, and Now{\'e}}]{van2013scalarized}
Kristof Van~Moffaert, Madalina~M Drugan, and Ann Now{\'e}. 2013.
\newblock Scalarized multi-objective reinforcement learning: Novel design techniques.
\newblock In \emph{2013 IEEE symposium on adaptive dynamic programming and reinforcement learning (ADPRL)}, pages 191--199. IEEE.

\bibitem[{Wang et~al.(2026{\natexlab{a}})Wang, Zheng, Zhang, Zhu, Cheng, Xie, Wang, and Feng}]{wang2026perm}
Chengbing Wang, Wuqiang Zheng, Yang Zhang, Fengbin Zhu, Junyi Cheng, Yi~Xie, Wenjie Wang, and Fuli Feng. 2026{\natexlab{a}}.
\newblock {PERM}: Psychology-grounded empathetic reward modeling for large language models.
\newblock In \emph{Findings of the Association for Computational Linguistics: ACL 2026}, pages 7350--7373.

\bibitem[{Wang et~al.(2025{\natexlab{a}})Wang, Shen, Yu, and Xia}]{wang2025flexible}
Fanfan Wang, Xiangqing Shen, Jianfei Yu, and Rui Xia. 2025{\natexlab{a}}.
\newblock Flexible thinking for multimodal emotional support conversation via reinforcement learning.
\newblock In \emph{Findings of the Association for Computational Linguistics: EMNLP 2025}, pages 1341--1356.

\bibitem[{Wang et~al.(2024{\natexlab{a}})Wang, Xu, Leong, Li, and Li}]{wang2024muffin}
Jiashuo Wang, Chunpu Xu, Chak~Tou Leong, Wenjie Li, and Jing Li. 2024{\natexlab{a}}.
\newblock Muffin: Mitigating unhelpfulness in emotional support conversations with multifaceted ai feedback.
\newblock In \emph{Findings of the Association for Computational Linguistics: ACL 2024}, pages 567--585.

\bibitem[{Wang et~al.(2026{\natexlab{b}})Wang, Ma, Zhang, Chen, He, Luo, Lv, Jiang, Xie, Wang, Li, Li, Ye, Li, Yang, Li, Tu, and Li}]{wang2026rlver}
Peisong Wang, Ruotian Ma, Bang Zhang, Xingyu Chen, Zhiwei He, Kang Luo, Qingsong Lv, Qingxuan Jiang, Zheng Xie, Shanyi Wang, Cixing Li, Yuan Li, Fanghua Ye, Jian Li, Yifan Yang, Jia Li, Zhaopeng Tu, and Xiaolong Li. 2026{\natexlab{b}}.
\newblock {RLVER}: Reinforcement learning with verifiable emotion rewards for empathetic agents.
\newblock In \emph{The Fourteenth International Conference on Learning Representations}.

\bibitem[{Wang et~al.(2020)Wang, Wei, Dong, Bao, Yang, and Zhou}]{wang2020minilm}
Wenhui Wang, Furu Wei, Li~Dong, Hangbo Bao, Nan Yang, and Ming Zhou. 2020.
\newblock Minilm: Deep self-attention distillation for task-agnostic compression of pre-trained transformers.
\newblock \emph{Advances in neural information processing systems}, 33:5776--5788.

\bibitem[{Wang et~al.(2025{\natexlab{b}})Wang, Zhao, Gu, Jiang, Chen, and Ji}]{wang2025straq}
Xiaoyu Wang, Yue Zhao, Qingqing Gu, Zhonglin Jiang, Yong Chen, and Luo Ji. 2025{\natexlab{b}}.
\newblock Convert language model into a value-based strategic planner.
\newblock In \emph{Proceedings of the 63rd Annual Meeting of the Association for Computational Linguistics (Volume 6: Industry Track)}.

\bibitem[{Wang et~al.(2024{\natexlab{b}})Wang, Yu, Yao, Zeng, Yang, Wang, Chen, Jiang, Xie, Wang, Xie, Ye, Zhang, and Zhang}]{wang2024pandalm}
Yidong Wang, Zhuohao Yu, Wenjin Yao, Zhengran Zeng, Linyi Yang, Cunxiang Wang, Hao Chen, Chaoya Jiang, Rui Xie, Jindong Wang, Xing Xie, Wei Ye, Shikun Zhang, and Yue Zhang. 2024{\natexlab{b}}.
\newblock {PandaLM}: An automatic evaluation benchmark for {LLM} instruction tuning optimization.
\newblock In \emph{The Twelfth International Conference on Learning Representations}.

\bibitem[{Wang et~al.(2025{\natexlab{c}})Wang, Liu, Zhao, Liao, and Nie}]{wang2025peer}
Yunxiao Wang, Meng Liu, Sicheng Zhao, Lizi Liao, and Liqiang Nie. 2025{\natexlab{c}}.
\newblock {PEER}: Unified process--outcome reinforcement learning for structured empathetic reasoning.
\newblock \emph{arXiv preprint arXiv:2508.09521}.

\bibitem[{Wei et~al.(2022)Wei, Wang, Schuurmans, Bosma, Ichter, Xia, Chi, Le, and Zhou}]{wei2022chain}
Jason Wei, Xuezhi Wang, Dale Schuurmans, Maarten Bosma, Brian Ichter, Fei Xia, Ed~H. Chi, Quoc~V. Le, and Denny Zhou. 2022.
\newblock Chain-of-thought prompting elicits reasoning in large language models.
\newblock In \emph{Advances in Neural Information Processing Systems}, volume~35, pages 24824--24837.

\bibitem[{Yu et~al.(2024)Yu, Yao, Zhang, He, Han, Cui, Hu, Liu, Zheng, Sun, and Chua}]{yu2024rlhf}
Tianyu Yu, Yuan Yao, Haoye Zhang, Taiwen He, Yifeng Han, Ganqu Cui, Jinyi Hu, Zhiyuan Liu, Hai-Tao Zheng, Maosong Sun, and Tat-Seng Chua. 2024.
\newblock {RLHF-V}: Towards trustworthy {MLLMs} via behavior alignment from fine-grained correctional human feedback.
\newblock In \emph{Proceedings of the IEEE/CVF Conference on Computer Vision and Pattern Recognition}, pages 13807--13816.

\bibitem[{Zhang et~al.(2025)Zhang, Meng, Luo, Han, Liao, Cambria, and Fei}]{zhang2025towards}
Han Zhang, Zixiang Meng, Meng Luo, Hong Han, Lizi Liao, Erik Cambria, and Hao Fei. 2025.
\newblock Towards multimodal empathetic response generation: A rich text-speech-vision avatar-based benchmark.
\newblock In \emph{Proceedings of the ACM on Web Conference 2025}, pages 2872--2881.

\bibitem[{Zhang et~al.(2023)Zhang, Naradowsky, and Miyao}]{zhang2023ask}
Qiang Zhang, Jason Naradowsky, and Yusuke Miyao. 2023.
\newblock Ask an expert: Leveraging language models to improve strategic reasoning in goal-oriented dialogue models.
\newblock In \emph{Findings of the Association for Computational Linguistics: ACL 2023}, pages 6665--6694.

\bibitem[{Zhang et~al.(2020)Zhang, Kishore, Wu, Weinberger, and Artzi}]{zhang2020bertscore}
Tianyi Zhang, Varsha Kishore, Felix Wu, Kilian~Q. Weinberger, and Yoav Artzi. 2020.
\newblock Bertscore: Evaluating text generation with bert.
\newblock In \emph{International Conference on Learning Representations}.

\bibitem[{Zhang et~al.(2024)Zhang, Kong, Wang, Sun, Wang, Feng, Wang, Zhang, and Song}]{zhang2024stickerconv}
Yiqun Zhang, Fanheng Kong, Peidong Wang, Shuang Sun, Lingshuai Wang, Shi Feng, Daling Wang, Yifei Zhang, and Kaisong Song. 2024.
\newblock {STICKERCONV}: Generating multimodal empathetic responses from scratch.
\newblock In \emph{Proceedings of the 62nd Annual Meeting of the Association for Computational Linguistics (Volume 1: Long Papers)}, pages 7707--7733. Association for Computational Linguistics.

\bibitem[{Zhao et~al.(2024)Zhao, Li, Chen, Kong, Wang, Huang, Gu, Wang, Wang, Liang, Li, Teng, Xiao, and Wang}]{zhao2024esc}
Haiquan Zhao, Lingyu Li, Shisong Chen, Shuqi Kong, Jiaan Wang, Kexin Huang, Tianle Gu, Yixu Wang, Jian Wang, Dandan Liang, Zhixu Li, Yan Teng, Yanghua Xiao, and Yingchun Wang. 2024.
\newblock {ESC}-eval: Evaluating emotion support conversations in large language models.
\newblock In \emph{Proceedings of the 2024 Conference on Empirical Methods in Natural Language Processing}, pages 15785--15810.

\bibitem[{Zheng et~al.(2023)Zheng, Chiang, Sheng, Zhuang, Wu, Zhuang, Lin, Li, Li, Xing, Zhang, Gonzalez, and Stoica}]{zheng2023judging}
Lianmin Zheng, Wei-Lin Chiang, Ying Sheng, Siyuan Zhuang, Zhanghao Wu, Yonghao Zhuang, Zi~Lin, Zhuohan Li, Dacheng Li, Eric~P. Xing, Hao Zhang, Joseph~E. Gonzalez, and Ion Stoica. 2023.
\newblock Judging {LLM}-as-a-judge with {MT}-bench and chatbot arena.
\newblock \emph{Advances in Neural Information Processing Systems}, 36:46595--46623.

\bibitem[{Zou et~al.(2026)Zou, Wang, Shen, Xiao, Ma, Li, Mao, and Cambria}]{zou2026affective}
Chenghui Zou, Ning Wang, Tiesunlong Shen, Luwei Xiao, Chuan Ma, Xiangpeng Li, Rui Mao, and Erik Cambria. 2026.
\newblock Affective flow language model for emotional support conversation.
\newblock \emph{arXiv preprint arXiv:2602.08826}.

\end{thebibliography}

\appendix

\section{Experimental Setup}
\label{sec:appendix_setup}

This section provides the core implementation details required to reproduce our experiments. We report the main model, training, decoding, evaluation, reward, and hardware settings, while omitting unchanged optimizer defaults such as AdamW betas and other non-critical implementation details.

\subsection{Training Configuration}
\label{sec:appendix_training_config}

Our policy is initialized through supervised fine-tuning, followed by a format-and-outcome GRPO warm-up and full \textsc{FACES} training with gated rewards and a clipped GRPO objective over fine-grained token-level advantages. Table~\ref{tab:core_training_hyperparams} reports the core \textsc{FACES} configuration, including its GRPO-specific hyperparameters.

DPO, SimPO, and RLOO use the same backbone, SFT initialization, training prompts, and evaluation protocol, while retaining their respective optimization objectives and update rules.

\begin{table}[!htbp]
\centering
\small
\renewcommand{\arraystretch}{1.08}
\begin{tabular}{@{}p{0.47\linewidth}p{0.43\linewidth}@{}}
\toprule
\textbf{Item} & \textbf{Value} \\
\midrule
Policy backbone & Qwen2.5-VL-7B-Instruct \\
Reward critic & Qwen3-VL-8B-Instruct \\
SFT learning rate & $5\times10^{-5}$ \\
GRPO learning rate & $1\times10^{-5}$ \\
Optimizer & AdamW \\
SFT / warm up / FACES epochs & 3 / 3 / 3 \\
Global batch size & 32 \\
Learning rate warmup ratio & 0.05 \\
GRPO group size $G$ & 4 \\
GRPO clip coefficient $\epsilon_{\mathrm{clip}}$ & 0.2 \\
Policy updates per rollout & 4 \\
KL coefficient $\beta$ & 0.001 \\
GRPO rollout decoding & Sampling with temperature 1.3 and top-$p$ 0.95 \\
Maximum rollout length & 192 tokens \\
Maximum sequence length & 1024 tokens \\
Maximum image side length & 384 pixels \\
Maximum video frames & 16 \\
LoRA rank / alpha / dropout & 8 / 32 / 0.05 \\
LoRA target modules & All linear layers \\
Precision & bfloat16 \\
Training hardware & 2 NVIDIA A6000 GPUs \\
\bottomrule
\end{tabular}
\caption{Core training configuration.}
\label{tab:core_training_hyperparams}
\end{table}

\FloatBarrier

\subsection{Inference and Evaluation Settings}
\label{sec:appendix_inference_eval}

The FIRES automatic scores in Table~\ref{tab:automatic_evaluation} are taken from \citet{wang2025flexible}. For local response-level evaluation, FIRES responses are generated using the inference settings in Table~\ref{tab:inference_eval_settings}, which also lists the primary Qwen and independent GLM judge configurations. Test-time generation is greedy, whereas GRPO rollouts sample at temperature 1.3 and top-$p$ 0.95 to produce diverse training candidates.

The primary Qwen judge also uses deterministic decoding with temperature 0 and sampling disabled. For cross-judge evaluation, GLM-4.6V-Flash evaluates exactly the same fixed candidate responses as the primary Qwen judge. Judge-specific output budgets affect only the evaluation outputs and do not alter the candidate responses. We additionally disable GLM thinking mode to reduce variation arising from judge-side reasoning.

Both judges receive the same dialogue context, visual evidence, length-controlled responses, and candidate ordering, with up to eight images. Bootstrap analyses use 5,000 resamples and seed 42. Judge prompts and controlled protocols appear in Appendices~\ref{sec:appendix_eval_prompts} and~\ref{sec:appendix_controlled_eval}.

For directly prompted MLLMs, metrics use complete responses and references. Distinct-3 uses corpus-level aggregation. BERTScore-F1 uses final-layer RoBERTa-large without IDF or baseline rescaling. DIFF-EPITOME uses mean squared error between candidate and reference scores.

\begin{table}[!htbp]
\centering
\small
\renewcommand{\arraystretch}{1.08}
\begin{tabular}{@{}p{0.47\linewidth}p{0.43\linewidth}@{}}
\toprule
\textbf{Item} & \textbf{Value} \\
\midrule
Primary evaluation judge & Qwen2.5-VL-32B-Instruct-AWQ \\
Cross-family evaluation judge & GLM-4.6V-Flash \\
\midrule
\multicolumn{2}{@{}l@{}}{\textit{Local generation and Qwen judge}} \\
Decoding strategy & Greedy decoding \\
Maximum new tokens & 256 \\
Temperature & 0 \\
Sampling & False \\
Top-$p$ & Not used \\
Maximum input images & 8 \\
\midrule
\multicolumn{2}{@{}l@{}}{\textit{GLM judge}} \\
Maximum output tokens & 768 \\
Temperature & 0 \\
Thinking mode & Disabled \\
Maximum input images & 8 \\
\midrule
\multicolumn{2}{@{}l@{}}{\textit{Statistical evaluation}} \\
Bootstrap samples & 5000 \\
Bootstrap seed & 42 \\
\bottomrule
\end{tabular}
\caption{Inference-time generation and automatic evaluation settings, with judge-specific configurations reported separately.}
\label{tab:inference_eval_settings}
\end{table}

\FloatBarrier

\subsection{Reward Configuration}
\label{sec:appendix_reward_config}

FACES uses a non-compensatory reward fusion strategy. All continuous reward components are normalized to $[0,1]$, while the format score $r_i^{\mathrm{Fmt}}$ is binary. We first aggregate visual grounding, emotion recognition, and situation understanding into a cognitive reward:
\begin{equation}
R_i^{\mathrm{Cog}} =
\frac{
r_i^{\mathrm{Vis}} +
r_i^{\mathrm{Emo}} +
r_i^{\mathrm{Sit}}
}{3}.
\end{equation}

The final reward is then computed with format validity as a hard gate, and strategy and outcome as multiplicative gates:
\begin{equation}
R_i =
\begin{cases}
0, & r_i^{\mathrm{Fmt}} = 0, \\
R_i^{\mathrm{Cog}} \cdot r_i^{\mathrm{Str}} \cdot r_i^{\mathrm{Out}}, & r_i^{\mathrm{Fmt}} = 1.
\end{cases}
\end{equation}

Table~\ref{tab:reward_components} summarizes validation-selected reward settings. The critic decodes deterministically with up to 512 new tokens. CLIP averages the four highest framewise cosine similarities, rescales the mean by $(1+s)/2$, then applies the 0.60 threshold. MiniLM uses masked mean pooling and L2-normalized embeddings. Parsing retains the first think/response blocks and last nonempty repeated stage. Missing visual/emotion inputs and invalid critic fields score zero; missing Situation/Strategy spans are passed as empty text.

\begin{table}[!htbp]
\centering
\scriptsize
\renewcommand{\arraystretch}{1.12}
\begin{tabular}{@{}p{0.30\linewidth}p{0.16\linewidth}p{0.44\linewidth}@{}}
\toprule
\textbf{Component} & \textbf{Range} & \textbf{Role in reward fusion} \\
\midrule
Format $r_i^{\mathrm{Fmt}}$ & $\{0,1\}$ & Hard gate in the full FACES reward stage; invalid outputs receive 0 in final reward fusion. \\
Visual $r_i^{\mathrm{Vis}}$ & $[0,1]$ & $1/3$ in cognitive reward \\
Emotion $r_i^{\mathrm{Emo}}$ & $[0,1]$ & $1/3$ in cognitive reward \\
Situation $r_i^{\mathrm{Sit}}$ & $[0,1]$ & $1/3$ in cognitive reward \\
Strategy $r_i^{\mathrm{Str}}$ & $[0,1]$ & Multiplicative gate \\
Outcome $r_i^{\mathrm{Out}}$ & $[0,1]$ & Multiplicative gate \\
Cognitive $R_i^{\mathrm{Cog}}$ & $[0,1]$ & Mean of visual, emotion, and situation \\
Final reward $R_i$ & $[0,1]$ & Product of cognitive, strategy, and outcome if format is valid \\
\bottomrule
\end{tabular}
\caption{Reward components and non-compensatory fusion rule.}
\label{tab:reward_components}
\end{table}

\subsection{Reward Reliability and Cost}
\label{sec:appendix_reward_reliability_cost}

For 1,829 format-valid offline-diagnostic candidates, MLLM-scored dimensions use absolute-agreement ICC(2,1) per prompt and ICC(2,3) for three-prompt means. The same three annotators in Appendix~\ref{sec:appendix_human_annotation_protocol} independently rate 150 candidates stratified by gated-reward quintile from 1 to 5 on five reward dimensions and Overall Quality. Table~\ref{tab:reward_reliability} reports Spearman's $\rho$ for each dimensional reward against its mean human score and for gated reward against mean human Overall Quality.

Three-prompt averaging raises ICC from 0.74--0.79 to 0.90--0.92; gated reward correlates with human Overall Quality at $\rho=0.72$, supporting its use in training.

To quantify reward computation cost, we benchmark 100 GRPO steps with the same Qwen2.5-VL-7B policy, batch and group sizes, generation limit, and two NVIDIA A6000 GPUs. FIRES uses Format, ROUGE-L, and DIFF-EPITOME/RoBERTa rewards; \textsc{FACES} uses CLIP, MiniLM, and the Qwen3-VL reward critic. Table~\ref{tab:training_cost} shows that \textsc{FACES} uses 2.45 times as many GPU hours as FIRES. Neither uses reward models at deployment; Table~\ref{tab:inference_cost} summarizes inference measurements for \textsc{FACES} and FIRES on identical hardware.

\noindent\begin{minipage}{\columnwidth}
\centering
\footnotesize
\setlength{\tabcolsep}{2pt}
\renewcommand{\arraystretch}{1.08}
\begin{tabular}{@{}L{0.42\columnwidth}C{0.17\columnwidth}C{0.17\columnwidth}C{0.18\columnwidth}@{}}
\toprule
\textbf{Dimension}
& \textbf{ICC(2,1)}
& \textbf{ICC(2,3)}
& \makecell{\textbf{Human}\\$\rho$} \\
\midrule
Visual & -- & -- & 0.61 \\
Emotion & -- & -- & 0.68 \\
Situation & 0.74 & 0.90 & 0.66 \\
Strategy & 0.79 & 0.92 & 0.71 \\
Outcome & 0.77 & 0.91 & 0.74 \\
Gated reward vs. Overall Quality & -- & -- & 0.72 \\
\bottomrule
\end{tabular}
\captionof{table}{Reliability of the training-time reward system. Visual and Emotion are deterministic CLIP and MiniLM scores, so prompt-level ICC is not applicable. The last column reports reward--human correlations, all significant with $p<0.001$. Higher values indicate stronger reliability or alignment.}
\label{tab:reward_reliability}
\end{minipage}

\par\smallskip

\noindent\begin{minipage}{\columnwidth}
\centering
\footnotesize
\setlength{\tabcolsep}{2pt}
\renewcommand{\arraystretch}{1.08}
\begin{tabular*}{\columnwidth}{@{\extracolsep{\fill}}L{0.28\columnwidth}@{\hspace{6pt}}C{0.24\columnwidth}C{0.15\columnwidth}C{0.24\columnwidth}@{}}
\toprule
\textbf{Pipeline}
& \textbf{Wall-time (h)}
& \textbf{GPU-h}
& \textbf{Relative cost} \\
\midrule
FIRES & 17.2 & 34.4 & $1.00\times$ \\
\textsc{FACES} & 42.2 & 84.4 & $2.45\times$ \\
\bottomrule
\end{tabular*}
\captionof{table}{Training cost over 100 stable GRPO steps. Lower values are better for all three measures.}
\label{tab:training_cost}
\end{minipage}

\par\smallskip

\noindent\begin{minipage}{\columnwidth}
\centering
\footnotesize
\setlength{\tabcolsep}{2pt}
\renewcommand{\arraystretch}{1.08}
\begin{tabular}{@{}L{0.30\columnwidth}C{0.32\columnwidth}C{0.32\columnwidth}@{}}
\toprule
\textbf{Metric} & \textbf{FACES} & \textbf{FIRES} \\
\midrule
Latency (s) $\downarrow$ & $15.828 \pm 2.820$ & $13.752 \pm 2.145$ \\
Output length & 184.7 tokens & 128.9 tokens \\
Throughput & 11.67 tokens/s & 9.37 tokens/s \\
Peak memory & 20.130 GiB & 20.130 GiB \\
\bottomrule
\end{tabular}
\captionof{table}{Deployment-time inference cost measured on 100 samples using one NVIDIA RTX A6000. Latency is reported per sample. Reward models are used only during training.}
\label{tab:inference_cost}
\end{minipage}

\section{Prompt Templates}
\label{sec:appendix_prompts}

This section reports the prompt templates used in \textsc{FACES}. The prompt design follows three principles. First, training-time reward construction and post-hoc evaluation use separate prompt families because they serve different purposes: the former provides optimization signals for the policy, whereas the latter measures the quality of fixed model outputs under independently specified evaluation criteria. Second, for every scoring task, we use three semantically equivalent prompt variants that preserve the same input fields, scoring criteria, score range, and output schema while varying their surface realization. This design reduces sensitivity to any single prompt formulation without changing the underlying evaluation objective. Third, all prompts require machine-readable JSON outputs, enabling deterministic parsing and consistent aggregation across instances.

For training-time reward scoring, Table~\ref{tab:training_reward_prompt} presents one representative template. For post-hoc MLLM evaluation, we similarly use three variants for each pointwise metric and pairwise preference setting. The shared evaluation rules and representative templates are reported in Table~\ref{tab:eval_common_rules}, Table~\ref{tab:pointwise_metrics}, Table~\ref{tab:pointwise_eval_prompt}, and Table~\ref{tab:pairwise_eval_prompt}. The complete prompt variants used in the experiments, together with the corresponding parsing and aggregation code, are available in the code repository.

\subsection{Training-time Reward Prompts}
\label{sec:appendix_training_prompts}

The training-time MLLM reward prompts are used to score situation understanding, support strategy, and response outcome during reward construction. Candidate reasoning and response spans, together with their corresponding reference annotations, are dynamically inserted through placeholders such as \textcolor{red}{\{context\}} and \textcolor{red}{\{candidate\_response\}}. The judge assesses semantic consistency, functional alignment, and contextual appropriateness, while penalizing unnecessary verbosity, and returns structured scores from 1 to 5 for the three dimensions. Visual perception and emotion understanding are evaluated separately using CLIP and MiniLM rather than the MLLM prompt. Scores obtained from the three prompt variants are normalized and then averaged before being used in reward fusion.
\begin{table*}[!t]
\centering
\small
\begin{tabular}{p{0.95\linewidth}}
\toprule
\textbf{Representative Training-time Reward Prompt} \\
\midrule
\setlength{\parindent}{0pt}
You are a strict evaluator for multimodal empathetic dialogue.

Evaluate the candidate using the dialogue context and the corresponding reference annotations. Your judgment must consider all of the following:

1. semantic consistency with the corresponding reference part;

2. functional alignment with the intended role of the reference part;

3. appropriateness to the current dialogue context and available evidence.

Score the candidate on the following three dimensions using integer scores from 1 to 5:

- Situation: whether the candidate situation inference is semantically consistent with the reference situation, functionally appropriate for understanding the user's condition, and grounded in the available evidence.

- Strategy: whether the candidate support strategy is semantically aligned with the reference strategy, functionally appropriate for the user's needs, and psychologically safe in the current context.

- Outcome: whether the final response is semantically and functionally aligned with the reference response, while remaining empathetic, helpful, and contextually appropriate.

Scoring rule:

- 5 = great

- 4 = good

- 3 = acceptable

- 2 = weak

- 1 = poor

- Penalize overly long, verbose, repetitive, or formulaic candidates with redundant or padded wording. If a candidate is substantially longer than the reference and the extra content does not improve semantic or functional alignment, lower the corresponding score(s).

[CONTEXT] \newline
\textcolor{red}{\{context\}}

[CANDIDATE\_SITUATION] \newline
\textcolor{red}{\{candidate\_situation\}}

[REFERENCE\_SITUATION] \newline
\textcolor{red}{\{reference\_situation\}}

[CANDIDATE\_STRATEGY] \newline
\textcolor{red}{\{candidate\_strategy\}}

[REFERENCE\_STRATEGY] \newline
\textcolor{red}{\{reference\_strategy\}}

[CANDIDATE\_RESPONSE] \newline
\textcolor{red}{\{candidate\_response\}}

[REFERENCE\_RESPONSE] \newline
\textcolor{red}{\{reference\_response\}}

Return JSON only in the following format: \newline
\{"situation\_score": 1, "strategy\_score": 1, "outcome\_score": 1\} \\
\bottomrule
\end{tabular}
\caption{Representative prompt template for training-time MLLM reward scoring. We use three semantically equivalent prompt variants in the actual reward-scoring pipeline.}
\label{tab:training_reward_prompt}
\end{table*}

\subsection{MLLM Evaluation Prompts}
\label{sec:appendix_eval_prompts}

For automatic MLLM evaluation, we use complementary pointwise scoring and blind pairwise preference comparison. Pointwise evaluation measures each response independently along Empathy, Multimodal Appropriateness, and Context--Response Consistency, whereas pairwise evaluation asks the judge to select the better overall emotional-support response or return a tie. The former provides interpretable dimension-level measurements, while the latter captures holistic quality when different strengths and weaknesses must be considered jointly. In pairwise comparisons, model identities are hidden, and both candidates are evaluated using the same dialogue context and visual evidence.

Unlike training-time reward scoring, post-hoc MLLM judges receive no reference situation, strategy, or response annotations. Consequently, a candidate can be assessed for contextual appropriateness without requiring its wording or support strategy to match a single dataset reference.

For each pointwise metric, we apply three semantically equivalent prompt variants and average the resulting scores. Pairwise evaluation uses the same number of prompt variants together with bidirectional candidate swapping and majority voting, reducing sensitivity to prompt wording and presentation order. For the length-controlled evaluations reported in the main results and Appendix~\ref{sec:appendix_controlled_eval}, all think-related fields and think-specific criteria are removed from the judge input and prompt. The judge therefore evaluates only the visible, length-controlled final responses rather than explicit reasoning traces. All prompts require structured JSON outputs; invalid or unparsable pairwise outputs are conservatively treated as ties. Table~\ref{tab:eval_common_rules} summarizes the shared evaluation rules, Table~\ref{tab:pointwise_metrics} defines the pointwise dimensions, and Tables~\ref{tab:pointwise_eval_prompt} and~\ref{tab:pairwise_eval_prompt} present the representative prompts.

\begin{table*}[!t]
\centering
\small
\renewcommand{\arraystretch}{1.12}
\begin{tabular}{p{0.25\linewidth}p{0.68\linewidth}}
\toprule
\textbf{Rule} & \textbf{Description} \\
\midrule
JSON-only output & The judge must return a valid JSON object and no extra explanation. \\
Score range & Pointwise scores are integers from 1 to 5, where 5 indicates the best quality. \\
Pairwise labels & Pairwise comparison returns one of \texttt{A}, \texttt{B}, or \texttt{TIE}. \\
Model anonymity & Candidate names are hidden from the judge and replaced with Candidate A and Candidate B. \\
Think section & Present in the base templates; removed in length-controlled variants. \\
Verbosity penalty & The judge is instructed to penalize overly long, repetitive, verbose, or formulaic responses when the extra content does not improve helpfulness, empathy, or contextual alignment. \\
Pointwise aggregation & Each response is scored from 1 to 5 under three prompt variants, and the three scores are averaged for each evaluation dimension. \\
Pairwise aggregation & For each of the three prompt variants, the judge is queried twice, once with FACES as Candidate A and FIRES as Candidate B, and once with the candidate order reversed. After mapping each A, B, or Tie judgment back to the corresponding model identity, the two order-specific judgments are combined as follows: agreement is retained; if one judgment is Tie, the other is retained; and conflicting non-tie judgments produce Tie. The resulting three variant-level decisions are aggregated by comparing the FACES and FIRES vote counts, with equal counts producing Tie. An invalid or unparsable output is treated as Tie for that candidate order, without retry or exclusion. \\
\bottomrule
\end{tabular}
\caption{Common rules for MLLM-based automatic evaluation.}
\label{tab:eval_common_rules}
\end{table*}

\begin{table*}[!t]
\centering
\small
\renewcommand{\arraystretch}{1.12}
\setlength{\tabcolsep}{4pt} 
\resizebox{0.99\linewidth}{!}{%
\begin{tabular}{@{}>{\raggedright\arraybackslash}p{0.25\linewidth}
                >{\raggedright\arraybackslash}p{0.51\linewidth}
                >{\raggedright\arraybackslash}p{0.23\linewidth}@{}}
\toprule
\textbf{Metric} & \textbf{Evaluation focus} & \textbf{JSON key} \\
\midrule
Empathy &
Whether the final response expresses appropriate empathy, emotional validation, and supportive wording. &
\parbox[t]{\linewidth}{\raggedright empathy\_text} \\

Multimodal appropriateness &
Whether the response is consistent with the dialogue context and available visual evidence. &
\parbox[t]{\linewidth}{\raggedright multimodal\_\\appropriateness} \\

Context-Response Cons. &
Whether the final response is consistent with the dialogue context, visual evidence, and support direction. &
\parbox[t]{\linewidth}{\raggedright context\_response\_\\consistency} \\
\bottomrule
\end{tabular}%
}
\caption{Pointwise MLLM evaluation metrics.}
\label{tab:pointwise_metrics}
\end{table*}

\begin{table*}[!t]
\centering
\small
\begin{tabular}{p{0.95\linewidth}}
\toprule
\textbf{Representative Pointwise Evaluation Prompt} \\
\midrule
\setlength{\parindent}{0pt}
You are an impartial evaluator for multimodal emotional support dialogue.

Evaluate the candidate output according to the specified metric. Use the dialogue context, any attached visual evidence, the model's think section if available, and the final response. If the think section is missing, it will be shown as \texttt{Not provided}.

Your judgment must consider:

1. whether the candidate is appropriate to the user's emotional state and dialogue context;

2. whether it is grounded in the available multimodal evidence;

3. whether it avoids unsupported, unsafe, overly verbose, repetitive, or formulaic content;

4. whether it satisfies the specific evaluation metric.

[EVALUATION\_METRIC] \newline
\textcolor{red}{\{metric\_name\}}

[METRIC\_DEFINITION] \newline
\textcolor{red}{\{metric\_definition\}}

[CONTEXT] \newline
\textcolor{red}{\{context\}}

[MODEL\_THINK] \newline
\textcolor{red}{\{think\}}

[CANDIDATE\_RESPONSE] \newline
\textcolor{red}{\{response\}}

Score the candidate from 1 to 5. Return JSON only in the following format: \newline
\textcolor{red}{\{json\_score\_format\}} \\
\bottomrule
\end{tabular}
\caption{Representative pointwise MLLM evaluation prompt. We use three semantically equivalent prompt variants for each pointwise metric.}
\label{tab:pointwise_eval_prompt}
\end{table*}

\begin{table*}[!t]
\centering
\small
\begin{tabular}{p{0.95\linewidth}}
\toprule
\textbf{Representative Pairwise Preference Prompt} \\
\midrule
\setlength{\parindent}{0pt}
You are an impartial judge for multimodal emotional support dialogue.

Compare Candidate A and Candidate B based on the dialogue context, any attached visual evidence, the provided think sections if available, and the final responses. The model identities are hidden. If a think section is missing, it will be shown as \texttt{Not provided}.

Choose the better candidate according to the following criteria:

1. empathy and emotional validation;

2. contextual relevance to the user's situation;

3. consistency with the visual evidence when available;

4. appropriateness and safety of the support strategy;

5. coherence between the think section and the final response;

6. avoidance of unnecessarily long, repetitive, verbose, or formulaic wording.

If both candidates are comparable in overall quality, choose \texttt{TIE}.

[CONTEXT] \newline
\textcolor{red}{\{context\}}

[CANDIDATE A THINK] \newline
\textcolor{red}{\{think\_a\}}

[CANDIDATE A RESPONSE] \newline
\textcolor{red}{\{response\_a\}}

[CANDIDATE B THINK] \newline
\textcolor{red}{\{think\_b\}}

[CANDIDATE B RESPONSE] \newline
\textcolor{red}{\{response\_b\}}

Return JSON only in the following format: \newline
\{"winner": "A"\}

The value of \texttt{winner} must be one of \texttt{A}, \texttt{B}, or \texttt{TIE}. \\
\bottomrule
\end{tabular}
\caption{Representative pairwise MLLM preference prompt. We use three semantically equivalent prompt variants, bidirectional candidate swapping, and majority voting to reduce prompt-order bias.}
\label{tab:pairwise_eval_prompt}
\end{table*}

\FloatBarrier

\clearpage

\section{Controlled Evaluation}
\label{sec:appendix_controlled_eval}

To examine whether the observed pairwise preference trend persists after restricting the information visible to the judge, we reuse the same underlying Qwen2.5-VL-32B-Instruct-AWQ judge model under controlled views. The overall evaluation setup remains unchanged, including three semantically equivalent prompt variants, bidirectional candidate swapping, and majority vote. Each controlled view exposes only the permitted candidate fields; when reasoning traces are hidden, the corresponding fields and evaluation criteria are removed from the prompt.

\subsection{Controlled Views}
\label{sec:appendix_controlled_views}

We consider a response-only view without a word limit and three length-controlled views. In the response-only no-cap view, the judge sees the dialogue context, attached images, and the final response only; the prompt excludes the think section and no think-specific evaluation criterion is used. This view shows the same preference trend, but it may still be affected by response-length differences. Therefore, we focus the reported controlled results on the length-controlled cap-30/cap-25/cap-15 views. In these views, each candidate response is prefix-truncated to at most $N$ words, where $N \in \{30,25,15\}$.

\subsection{Controlled Pairwise Results}
\label{sec:appendix_controlled_results}

Table~\ref{tab:controlled_pairwise_results} reports blind pairwise preference outcomes under the three length-controlled views (cap-30, cap-25, and cap-15) on the MESC-FT test set and MELD-ESC300.

\begin{table}[!htbp]
\centering
\small
\renewcommand{\arraystretch}{1.04}
\setlength{\tabcolsep}{4.5pt}
\begin{tabular}{@{}ccccc@{}}
\toprule
\textbf{Dataset} & \textbf{View} & \textbf{A Win} & \textbf{B Win} & \textbf{Tie} \\
\midrule
MESC-FT test set & cap-30 & 426 & 11 & 13 \\
MESC-FT test set & cap-25 & 411 & 18 & 21 \\
MESC-FT test set & cap-15 & 361 & 53 & 36 \\
MELD-ESC300 & cap-30 & 287 & 7 & 6 \\
MELD-ESC300 & cap-25 & 279 & 11 & 10 \\
MELD-ESC300 & cap-15 & 186 & 29 & 85 \\
\bottomrule
\end{tabular}
\caption{Controlled pairwise evaluation results. Each cell reports majority-vote decisions. Model A denotes Ours (Full), and Model B denotes FIRES (Baseline).}
\label{tab:controlled_pairwise_results}
\end{table}

Across both datasets, Model A remains consistently preferred under all length-controlled views. As the word cap becomes stricter, the win rate of Model A gradually decreases, but Model A still clearly outperforms Model B even under the most restrictive cap-15 setting. Since cap-15 removes explicit reasoning content and imposes the strongest constraint on visible response length, it provides the most conservative evidence in this appendix.

\subsection{Statistical Significance}
\label{sec:appendix_significance}

To assess whether the observed preference advantages in the controlled pairwise evaluation are statistically reliable, we further perform significance testing on the main comparisons. For each dataset and each reported length-controlled view, we apply a two-sided exact binomial sign test over decisive (i.e., non-tie) comparisons only, and estimate 95\% confidence intervals for the FACES preference share via bootstrap resampling. This analysis does not introduce any new evaluation criterion; rather, it provides a statistical summary, including confidence intervals, of the majority-vote pairwise results reported in the main text and appendix.

The results show that FACES remains significantly preferred over the baseline under all reported length-controlled views on both the MESC-FT test set and MELD-ESC300. Notably, even under the most conservative cap-15 condition, where explicit reasoning traces are hidden and the judge can access only length-truncated final responses, FACES still attains a decisive preference share of 87.2\% on the MESC-FT test set (95\% CI: 83.9\%--90.4\%, $p \approx 2.0 \times 10^{-57}$) and 86.5\% on MELD-ESC300 (95\% CI: 81.9\%--91.2\%, $p \approx 3.1 \times 10^{-29}$). These results suggest that the observed preference gains are not merely driven by visible CoT content or longer responses, but remain robust under substantially restricted judging conditions.

\subsection{Cross-Judge Robustness}
\label{sec:appendix_cross_judge}

To examine whether the preference results are specific to a Qwen-family evaluator, we add GLM-4.6V-Flash as an independent judge. We retain the same three prompt variants, bidirectional candidate swapping, length control, and majority voting. Each result reports FACES win/FIRES win/tie.

\begin{table*}[t]
\centering
\scriptsize
\setlength{\tabcolsep}{4pt}
\begin{tabular*}{\textwidth}{@{\extracolsep{\fill}}llcllp{0.21\textwidth}}
\toprule
\textbf{Dataset} & \textbf{View} & \textbf{$n$} & \textbf{Qwen2.5-VL-32B} & \textbf{GLM-4.6V-Flash} & \textbf{GLM FACES Share Excl. Ties} \\
\midrule
MESC-FT & cap-15 & 450 & 80.22/11.78/8.00 & 84.9/6.0/9.1 & 93.4\% ($p=2.0{\times}10^{-81}$) \\
MESC-FT & cap-25 & 450 & 91.3/4.0/4.7 & 97.1/1.3/1.6 & -- \\
MESC-FT & response-only & 450 & 100.0/0.0/0.0 & 99.6/0.2/0.2 & -- \\
StickerConv & cap-15 & 300 & 63.7/18.3/18.0 & 64.3/19.4/16.3 & 76.8\% ($p=4.0{\times}10^{-18}$) \\
AvaMERG & cap-15 & 300 & 79.3/11.3/9.4 & 65.7/16.0/18.3 & 80.4\% ($p=1.3{\times}10^{-22}$) \\
\bottomrule
\end{tabular*}
\caption{Cross-judge pairwise preferences (\%). Both judges evaluate the same candidate pairs. Cap-$N$ views truncate final responses to at most $N$ words; response-only uses no word cap.}
\label{tab:cross_judge_robustness}
\end{table*}

\begin{table*}[t]
\centering
\scriptsize
\setlength{\tabcolsep}{4pt}
\begin{tabular*}{\textwidth}{@{\extracolsep{\fill}}llcll}
\toprule
\textbf{Dataset} & \textbf{Opponent} & \textbf{$n$} & \textbf{Qwen2.5-VL-32B} & \textbf{GLM-4.6V-Flash} \\
\midrule
MESC-FT & GPT-4o & 450 & 72.0/16.0/12.0 & 69.8/17.8/12.4 \\
MESC-FT & Claude-Haiku-4.5 & 450 & 74.0/14.4/11.6 & 71.3/16.3/12.4 \\
MESC-FT & Gemini-2.5-Flash-Lite & 450 & 72.7/15.3/12.0 & 70.2/17.1/12.7 \\
MESC-FT & GLM-4.1V-Thinking-FlashX & 450 & 76.0/12.7/11.3 & 72.9/14.7/12.4 \\
MESC-FT & Doubao-1.6-Vision & 450 & 76.7/12.0/11.3 & 74.0/13.8/12.2 \\
MESC-FT & Kimi-K2.5 & 450 & 71.6/16.2/12.2 & 69.3/18.0/12.7 \\
MELD-ESC300 & GPT-4o & 300 & 66.0/14.0/20.0 & 63.0/16.7/20.3 \\
MELD-ESC300 & Claude-Haiku-4.5 & 300 & 68.0/12.3/19.7 & 65.0/14.7/20.3 \\
MELD-ESC300 & Gemini-2.5-Flash-Lite & 300 & 67.0/13.3/19.7 & 64.0/15.7/20.3 \\
MELD-ESC300 & GLM-4.1V-Thinking-FlashX & 300 & 69.3/11.0/19.7 & 65.7/14.0/20.3 \\
MELD-ESC300 & Doubao-1.6-Vision & 300 & 70.0/10.7/19.3 & 67.0/13.0/20.0 \\
MELD-ESC300 & Kimi-K2.5 & 300 & 65.0/15.0/20.0 & 62.0/17.7/20.3 \\
\bottomrule
\end{tabular*}
\caption{Cap-15 blind pairwise comparisons between \textsc{FACES} and directly prompted strong MLLMs. Each cell reports FACES win/opponent win/tie percentages.}
\label{tab:strong_mllm_pairwise}
\end{table*}

Table~\ref{tab:cross_judge_robustness} reports the results obtained with the independent GLM judge. On MESC-FT, GLM assigns \textsc{FACES} win rates of 84.9\%, 97.1\%, and 99.6\% under cap-15, cap-25, and unrestricted response-only evaluation, respectively. Under cap-15, the corresponding win rates are 64.3\% on StickerConv and 65.7\% on AvaMERG. After ties are excluded, the cap-15 preference shares on MESC-FT, StickerConv, and AvaMERG are 93.4\%, 76.8\%, and 80.4\%, respectively, all significantly above chance under the reported sign tests ($p<0.001$). These results show a consistent length-controlled preference across the two judge families.

\subsection{Strong Direct-Prompted MLLMs}
\label{sec:appendix_strong_mllm_pairwise}

We compare \textsc{FACES} with six directly prompted strong MLLMs on MESC-FT (450 test instances) and MELD-ESC300 (300 external test instances). MESC-FT assesses in-domain performance, while MELD-ESC300 assesses cross-dataset transfer without additional fine-tuning on the external dataset. None of the six comparison models is task-fine-tuned, and each receives the same dialogue context, visual evidence, and response instruction.

For both datasets, we apply the cap-15 response-only blind pairwise protocol, removing reasoning traces and prefix-truncating each candidate response to at most 15 words. We use three semantically equivalent judge prompts, bidirectional candidate swapping to reduce presentation order bias, and majority voting. The same fixed candidate responses are evaluated by Qwen2.5-VL-32B-Instruct-AWQ and GLM-4.6V-Flash.

Table~\ref{tab:strong_mllm_pairwise} shows that \textsc{FACES} is preferred over all six opponents by both judges on both datasets. On MESC-FT, its win rates range from 71.6\% to 76.7\% under Qwen and from 69.3\% to 74.0\% under GLM. On MELD-ESC300, the corresponding ranges are 65.0\%--70.0\% and 62.0\%--67.0\%. The lower win rates on MELD-ESC300 indicate a smaller preference advantage under cross-dataset evaluation, but the direction of preference remains consistent for every opponent. GLM also assigns lower FACES win rates than Qwen throughout, while preserving the same overall conclusion. This consistency provides evidence that the observed advantage is not confined to evaluation by a Qwen-family judge.

Human evaluation provides complementary evidence for the comparison with GPT-4o. Among the six directly prompted MLLMs, only GPT-4o is included in this human comparison. Table~\ref{tab:human_gpt4o} reports the complete win, loss, and tie counts under the cap-15 blind protocol. Human annotators prefer \textsc{FACES} in 342 of 450 MESC-FT instances (76.0\%) and 216 of 300 MELD-ESC300 instances (72.0\%). These preferences agree in direction with both MLLM judges on both datasets. The human results support the comparison with GPT-4o specifically; comparisons with the other five MLLMs rely on the two automatic judges.

\begin{table}[t]
\centering
\small
\setlength{\tabcolsep}{3pt}
\begin{tabular}{lccc}
\toprule
\textbf{Dataset} & \textbf{FACES Win} & \textbf{GPT-4o Win} & \textbf{Tie} \\
\midrule
MESC-FT & 342 (76.0\%) & 63 (14.0\%) & 45 (10.0\%) \\
MELD-ESC300 & 216 (72.0\%) & 40 (13.3\%) & 44 (14.7\%) \\
\bottomrule
\end{tabular}
\caption{Human pairwise preference between \textsc{FACES} and GPT-4o under the cap-15 blind protocol.}
\label{tab:human_gpt4o}
\end{table}

\section{Detailed Preference Evaluation}
\label{sec:appendix_preference_eval}

\subsection{Human Annotation Protocol}
\label{sec:appendix_human_annotation_protocol}

\begin{table*}[t]
\centering
\small
\setlength{\tabcolsep}{6pt}
\begin{tabular*}{\textwidth}{@{\extracolsep{\fill}}L{0.20\textwidth}L{0.72\textwidth}}
\toprule
\textbf{Criterion} & \textbf{Guideline} \\
\midrule
Empathy &
Whether the response recognizes and appropriately addresses the user's emotion. \\
Multimodal appropriateness &
Whether the response is consistent with the dialogue context and visual evidence. \\
Strategy suitability &
Whether the support strategy fits the user's current needs without evident risk. \\
Helpfulness &
Whether the response provides practical supportive value. \\
Tie rule &
Select a winner only when one response is clearly better overall. Select Tie when neither response has a clear overall advantage, including cases where the responses have different strengths and weaknesses. \\
\bottomrule
\end{tabular*}
\caption{Criteria and decision rule for human pairwise preference annotation.}
\label{tab:human_annotation_guidelines}
\end{table*}

The three primary annotators comprised two PhD students with experience in affective computing or emotional-support modeling and one lecturer in a related field. Each annotator evaluated all 450 MESC-FT test instances and all 300 MELD-ESC300 instances for the FACES--FIRES comparison. The additional FACES--GPT-4o comparison used the same annotators, complete evaluation sets, and annotation procedure; its results are reported in Appendix~\ref{sec:appendix_strong_mllm_pairwise}.

Each item presented the dialogue context, visual evidence, and two anonymized candidate responses under the cap-15 view defined in Appendix~\ref{sec:appendix_controlled_views}. Candidate identities were hidden and their presentation order was randomized. Annotators selected Candidate A, Candidate B, or Tie based on the criteria in Table~\ref{tab:human_annotation_guidelines}. The four criteria jointly guided a single overall preference decision rather than separate numerical scores for each criterion.

The three annotators evaluated each item independently without seeing one another's labels. A label supported by at least two annotators became the final preference label. When all three labels differed, a fourth externally recruited psychology expert adjudicated the case. All reported preference counts include these adjudicated decisions where applicable. Fleiss' $\kappa$ was computed from the three primary annotators' raw labels before adjudication, measuring agreement among human annotators.

The primary annotators were recruited internally at their university and compensated according to the actual annotation workload. The fourth psychology expert was externally recruited and compensated for adjudication.

\subsection{Human Preference Counts}
\label{sec:appendix_human_preference}

Tables~\ref{tab:human_preference_mesc} and \ref{tab:human_preference_meld} report the final human preference counts for \textsc{FACES} versus FIRES. Percentages use all evaluated instances as the denominator, including ties. The reported Fleiss' $\kappa$ values measure agreement before adjudication, following the procedure in Appendix~\ref{sec:appendix_human_annotation_protocol}.

\begin{table}[t]
\centering
{\fontsize{8pt}{9pt}\selectfont
\setlength{\tabcolsep}{2pt}
\renewcommand{\arraystretch}{0.98}
\begin{tabular*}{0.95\columnwidth}{@{\extracolsep{\fill}}ccccc}
\toprule
\textbf{Dataset} & \textbf{FACES Win} & \textbf{FIRES Win} & \textbf{Tie} & $\boldsymbol{\kappa}$ \\
\midrule
MESC-FT & \textbf{339 (75.3\%)} & 63 (14.0\%) & 48 (10.7\%) & 0.62 \\
\bottomrule
\end{tabular*}
}
\caption{Final human preferences between \textsc{FACES} and FIRES on the 450-instance MESC-FT test set. $\kappa$ denotes Fleiss' kappa.}
\label{tab:human_preference_mesc}
\end{table}

\begin{table}[t]
\centering
{\fontsize{8pt}{9pt}\selectfont
\setlength{\tabcolsep}{2pt}
\renewcommand{\arraystretch}{0.98}
\begin{tabular*}{0.95\columnwidth}{@{\extracolsep{\fill}}ccccc}
\toprule
\textbf{Dataset} & \textbf{FACES Win} & \textbf{FIRES Win} & \textbf{Tie} & $\boldsymbol{\kappa}$ \\
\midrule
MELD-ESC300 & \textbf{211 (70.3\%)} & 44 (14.7\%) & 45 (15.0\%) & 0.58 \\
\bottomrule
\end{tabular*}
}
\caption{Final human preferences between \textsc{FACES} and FIRES on the 300-instance MELD-ESC300 benchmark. $\kappa$ denotes Fleiss' kappa.}
\label{tab:human_preference_meld}
\end{table}

\begin{table*}[t]
\centering
\small
\setlength{\tabcolsep}{6pt}
\begin{tabular*}{\textwidth}{@{\extracolsep{\fill}}lccccc}
\toprule
\textbf{Dataset} & \textbf{$n$} & \textbf{Human F/FIRES/Tie} & \textbf{MLLM F/FIRES/Tie} & \textbf{Exact Agreement} & \textbf{Cohen's $\kappa$} \\
\midrule
MESC-FT & 450 & 339/63/48 & 361/53/36 & 74.7\% (336/450) & 0.316 \\
MELD-ESC300 & 300 & 211/44/45 & 186/29/85 & 71.7\% (215/300) & 0.441 \\
\bottomrule
\end{tabular*}
\caption{Agreement between final human preferences and Qwen2.5-VL-32B-Instruct-AWQ majority preferences for the FACES--FIRES comparison under cap-15. F denotes \textsc{FACES}; preference distributions are reported as counts.}
\label{tab:human_mllm_agreement}
\end{table*}

\begin{table*}[t]
\centering
\small
\setlength{\tabcolsep}{8pt}
\begin{tabular}{lcccccc}
\toprule
\textbf{Model} & \textbf{B-2 $\uparrow$} & \textbf{R-L $\uparrow$} & \textbf{M. $\uparrow$} & \textbf{F$_{\text{BERT}}$ $\uparrow$} & \textbf{Dist-3 $\uparrow$} & \textbf{DIFF-E. $\downarrow$} \\
\midrule
w/o format              & \textbf{10.91} & \textbf{21.84} & 15.15 & 85.03 & 80.47 & 0.4293 \\
w/o visual              & 8.89  & 20.59 & 19.92 & 86.51 & 98.74 & 0.4073 \\
w/o emotion             & 9.44  & 19.89 & 19.37 & 86.95 & 97.92 & 0.6108 \\
w/o situation           & 9.97  & 19.62 & 19.04 & 86.24 & 97.74 & 0.5918 \\
w/o strategy            & 9.23  & 20.27 & 17.92 & 84.86 & 97.05 & 0.5251 \\
Ours (Full)             & 9.68  & 21.09 & \textbf{20.62} & \textbf{87.14} & \textbf{99.63} & \textbf{0.3643} \\
\bottomrule
\end{tabular}
\caption{Additional dimension-level ablation results on the MESC-FT test set. All metrics are higher-is-better except DIFF-E.}
\label{tab:additional_ablation}
\end{table*}

\subsection{Human--MLLM Agreement}
\label{sec:appendix_human_mllm_agreement}

We assess sample-level agreement for the FACES--FIRES comparison by pairing the final human preference label with the majority-vote label from Qwen2.5-VL-32B-Instruct-AWQ on the same cap-15 candidate pair. Both labels take one of three nominal values: FACES win, FIRES win, or Tie. We retain ties as a category and report exact agreement and unweighted Cohen's $\kappa$ in Table~\ref{tab:human_mllm_agreement}.

Exact agreement is the proportion of instances with identical human and MLLM labels. Cohen's $\kappa$ additionally adjusts for agreement expected from their marginal label distributions:

\begin{align}
\kappa
&=
\frac{p_o-p_e}{1-p_e},
\label{eq:human_mllm_kappa}\\
p_e
&=
\sum_{c\in\mathcal{C}}p_H(c)p_M(c),
\label{eq:expected_agreement}
\end{align}

where $p_o$ is the observed exact agreement rate,
$\mathcal{C}=\{\text{FACES win},\text{FIRES win},\text{Tie}\}$,
and $p_H(c)$ and $p_M(c)$ are the human and MLLM marginal label proportions.

MESC-FT has higher exact agreement but lower Cohen's $\kappa$ because both label distributions are more concentrated on \textsc{FACES}. The expected chance agreement is approximately 0.629 on MESC-FT, compared with 0.493 on MELD-ESC300. Both datasets therefore show agreement beyond chance, although substantial sample-level disagreement remains. These results support consistency with human preferences without establishing that the MLLM judge can replace human evaluation.

This analysis concerns the post-hoc Qwen judge on FACES--FIRES pairs. It does not measure agreement with GLM or on FACES--GPT-4o pairs. The separate analysis of training rewards against human numerical scores is reported in Appendix~\ref{sec:appendix_reward_reliability_cost}.

\begin{figure}[t]
    \centering
    \includegraphics[width=\columnwidth]{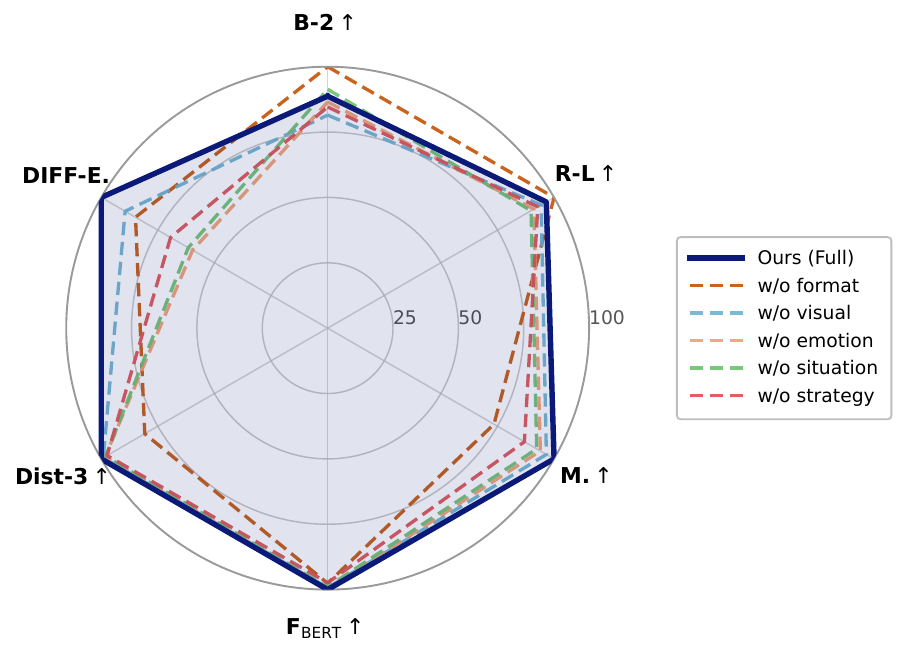}
    \caption{Normalized radar chart of dimension-level ablations. Larger values indicate better performance; DIFF-E. is reversed. FACES remains most balanced across metrics.}
    \label{fig:appendix_ablation_radar}
\end{figure}

\begin{table*}[t]
\centering
\small
\setlength{\tabcolsep}{5pt}
\begin{tabular*}{\textwidth}{@{\extracolsep{\fill}}L{0.16\textwidth}L{0.29\textwidth}L{0.25\textwidth}L{0.20\textwidth}}
\toprule
\textbf{Dimension} & \textbf{Targeted failure mode} & \textbf{Responsible span} & \textbf{Scoring signal} \\
\midrule
Visual & Incorrect or missing use of visual evidence & Visual-perception span & CLIP similarity \\
Emotion & Missing or incorrect affect recognition & Emotion-understanding span & MiniLM similarity \\
Situation & Incorrect attribution of the user's condition & Situation-understanding span & MLLM judgment \\
Strategy & Inappropriate or unsafe support strategy & Strategy-planning span & MLLM judgment; multiplicative gate \\
Outcome & Unhelpful or misaligned final response & Final-response span & MLLM judgment; multiplicative gate \\
\bottomrule
\end{tabular*}
\caption{Roles of the five reward dimensions. The taxonomy follows the MESC-FT/FIRES flexible-thinking structure and adds Outcome to capture final-response failures.}
\label{tab:reward_dimension_roles}
\end{table*}

\section{Additional Ablation Analysis}
\label{sec:appendix_additional_ablation}

Table~\ref{tab:additional_ablation} reports ablations of individual reward dimensions and format control on MESC-FT. We ablate the dimensions used for intermediate reasoning and structural control, but do not ablate the outcome reward because it provides the shared final-response supervision needed for comparable training. Table~\ref{tab:reward_dimension_roles} summarizes the function, responsible semantic span, and scoring signal of each reward dimension. The normalized trends are shown in Figure~\ref{fig:appendix_ablation_radar}. These results complement the contribution-level ablations in the main text by showing how the ablated reward dimensions and format control affect performance across the six automatic evaluation metrics.

Ours (Full) achieves the best METEOR, BERTScore-F1, Distinct-3, and DIFF-EPITOME, indicating the most balanced overall profile. Removing the format constraint increases BLEU-2 and ROUGE-L, but substantially hurts semantic quality and diversity, suggesting that lexical overlap alone does not reflect better support quality. Removing emotion or situation rewards leads to larger DIFF-EPITOME degradation, while removing strategy reward lowers semantic metrics, showing that these dimensions provide complementary supervision for strategy-aligned responses. \subsection{Reward Aggregation Diagnostics}
\label{sec:appendix_reward_aggregation}

We compare four aggregation rules using the same candidate responses from 369 prompts, with five responses initially sampled per prompt. Excluding responses with format scores below 0.5 leaves 1,829 candidates. Each rule selects the 457 candidates with the highest rewards, approximately 25\% of the retained set. A critical failure is defined as a candidate whose Strategy or Outcome score falls in the bottom quartile of the corresponding score distribution across all retained candidates.

\begin{table*}[t]
\centering
\small
\setlength{\tabcolsep}{5pt}
\begin{tabular*}{\textwidth}{@{\extracolsep{\fill}}L{0.16\textwidth}L{0.37\textwidth}ccc}
\toprule
\textbf{Aggregation} & \textbf{Definition} & \textbf{Top-Str. $\uparrow$} & \textbf{Top-Out. $\uparrow$} & \textbf{Critical-Fail $\downarrow$} \\
\midrule
Linear & Equal-weight mean over five dimensions & 0.325 & 0.328 & 2.4\% \\
Geometric mean & Geometric mean over five dimensions & 0.313 & 0.317 & 0.0\% \\
Tchebycheff & Minimum over five dimensions & 0.314 & 0.311 & 0.0\% \\
Gated (Ours) & mean(Visual, Emotion, Situation) $\times$ Strategy $\times$ Outcome & \textbf{0.343} & \textbf{0.346} & \textbf{0.0\%} \\
\bottomrule
\end{tabular*}
\caption{Offline comparison of reward aggregation schemes. Top-Str. and Top-Out. denote mean Strategy and Outcome scores among top-25\% candidates.}
\label{tab:aggregation_diagnostic}
\end{table*}

\begin{table}[!t]
\centering
\small
\setlength{\tabcolsep}{2.5pt}
\renewcommand{\arraystretch}{1.08}
\begin{tabular*}{\columnwidth}{@{\extracolsep{\fill}}lccc@{}}
\toprule
\textbf{Diagnostic} & \textbf{Value} & \textbf{Linear} &
\makecell[c]{\textbf{Gated}\\\textbf{(Ours)}} \\
\midrule
High-risk region fraction & 65.0\% & -- & -- \\
Mean reward in $\mathcal{Q}_{\mathrm{HL}}$ & -- & 0.4465 & 0.0096 \\
Global mean reward & -- & 0.4320 & 0.0349 \\
$\mathcal{Q}_{\mathrm{HL}}$ / global mean & -- & 103.4\% & 27.6\% \\
\makecell[l]{Groups assigning maximum\\advantage to a critical failure}
& -- & 14.9\% & 11.1\% \\
\bottomrule
\end{tabular*}
\caption{Reward-geometry diagnostics for candidates with strong cognitive signals but weak Strategy or Outcome quality.}
\label{tab:reward_geometry}
\end{table}

Table~\ref{tab:aggregation_diagnostic} shows that geometric mean, Tchebycheff, and gated aggregation each yield a critical failure rate of 0.0\%, compared with 2.4\% for linear aggregation. Among the four methods, gated aggregation achieves the highest mean Strategy and Outcome scores. Because none of the candidates ranked in the top quartile by the other three aggregation rules meets the critical failure criterion, our additional diagnostic focuses on the 11 candidates that meet this criterion under linear aggregation. We compare these candidates with the gated top quartile using three signals that are not produced by the Qwen judge. Relative to the mean scores of the gated top quartile, these linear critical failures score 9.4\% lower on MiniLM similarity, 2.9\% lower on EPITOME, and 6.2\% lower on ROUGE-L. This diagnostic provides additional evidence that linear aggregation can assign high rewards to candidates that appear weaker on average under signals independent of the Qwen judge.

We further examine the reward geometry in the high-risk region

\begin{equation}
\mathcal{Q}_{\mathrm{HL}}
=
\left\{
R^{\mathrm{Cog}}\geq 0.6,\ 
r^{\mathrm{Str}}r^{\mathrm{Out}}<0.1
\right\},
\label{eq:high_risk_region}
\end{equation}

which contains 65.0\% of the format-valid candidates. This region represents candidates with strong cognitive signals but critically weak strategy or outcome quality.

Table~\ref{tab:reward_geometry} shows that the mean reward within $\mathcal{Q}_{\mathrm{HL}}$ is 103.4\% of the global mean under linear aggregation, compared with 27.6\% under gated aggregation. Across the 369 prompt groups, the proportion of groups in which a critical failure receives the maximum advantage decreases from 14.9\% under linear aggregation to 11.1\% under gated aggregation. These results show that gated aggregation suppresses rewards for candidates with strong cognitive signals but weak Strategy or Outcome quality. It also reduces, but does not eliminate, the selection of critical failures within prompt groups.

\subsection{Fine-grained Credit Diagnostic}
\label{sec:appendix_segadv_diagnostic}

To further examine span level credit assignment, we use a supplementary five dimensional diagnostic on 100 fixed MESC-FT validation instances sampled with seed 42. For each response, CLIP scores Visual consistency, MiniLM scores Emotion alignment, and Qwen2.5-VL-32B-Instruct-AWQ scores Situation, Strategy, and Response. Let $q_i^d$ denote the normalized score assigned to response $i$ on dimension $d$. For $d=\mathrm{Vis}$ and $d=\mathrm{Emo}$, $q_i^d$ is the corresponding CLIP or MiniLM score. For each Qwen evaluated dimension $d\in\{\mathrm{Sit},\mathrm{Str},\mathrm{Resp}\}$, the judge produces $s_{i,k}^{d}\in\{1,2,3,4,5\}$ under prompt variant $k$. We map these scores to $[0,1]$ and average them across the three variants:

\begin{equation}
q_i^d
=
\frac{1}{3}
\sum_{k=1}^{3}
\frac{s_{i,k}^{d}-1}{4}.
\label{eq:diagnostic_score}
\end{equation}

All five scores are in $[0,1]$, and the composite score is

\begin{equation}
\begin{aligned}
C_i={}&15q_i^{\mathrm{Vis}}
+20q_i^{\mathrm{Emo}}
+20q_i^{\mathrm{Sit}}\\
&+20q_i^{\mathrm{Str}}
+25q_i^{\mathrm{Resp}},
\end{aligned}
\label{eq:diagnostic_composite}
\end{equation}

which ranges from 0 to 100. This supplementary diagnostic differs from the three dimensional 1--5 MLLM evaluation used for the main results.

\begin{table}[t]
\centering
\small
\setlength{\tabcolsep}{4pt}
\begin{tabular}{@{}lcc@{}}
\toprule
\textbf{Variant} & \textbf{Composite $\uparrow$} & \textbf{Change} \\
\midrule
Full model & 50.6 & -- \\
Without span alignment & 45.2 & $-10.7\%$ \\
\bottomrule
\end{tabular}
\caption{Supplementary diagnostic for credit assignment to semantic spans.}
\label{tab:segadv_diagnostic}
\end{table}

Table~\ref{tab:segadv_diagnostic} shows a 10.7\% decrease in the composite score, dropping from 50.6 to 45.2, when feedback is decoupled from its corresponding spans, suggesting that broadcasting feedback across the entire generation hinders the targeted learning of critical empathetic behaviors.

\section{Hyperparameter Sensitivity Analysis}
\label{sec:appendix_sensitivity}

We vary $\tau_v$, $\gamma$, $m$, and $\lambda$ individually to assess FACES sensitivity. These set visual reward sparsity, the minimum scale for positive advantages, the number of final tokens receiving full advantage, and the weight of earlier tokens, respectively. Appendix~\ref{sec:appendix_setup} gives the fixed training, decoding, and evaluation settings. Defaults and sensitivity results use the MESC-FT validation set; we compare endpoint anchoring with linear and exponential decay using validation rewards and MESC-FT test-set response evaluations.

Table~\ref{tab:tau_sensitivity} shows moderate robustness to changes in the visual threshold $\tau_v$. Lower thresholds make visual rewards more permissive and slightly improve lexical overlap but weaken semantic discrimination; higher thresholds make rewards sparse and reduce semantic and MLLM performance. The default $\tau_v=0.60$ gives the best DIFF-E. and MLLM average, though not every metric.

\begin{table*}[p]
\centering
\begin{minipage}[c][0.92\textheight][s]{\textwidth}
\centering

\begingroup
\small
\setlength{\tabcolsep}{7pt}
\renewcommand{\arraystretch}{1.08}
\begin{tabular*}{\textwidth}{@{\extracolsep{\fill}}cccccccc@{}}
\toprule
$\boldsymbol{\tau_v}$ &
\textbf{B-2 $\uparrow$} &
\textbf{R-L $\uparrow$} &
\textbf{M. $\uparrow$} &
\textbf{F$_{\text{BERT}}$ $\uparrow$} &
\textbf{Dist-3 $\uparrow$} &
\textbf{DIFF-E. $\downarrow$} &
\textbf{MLLM Avg. $\uparrow$} \\
\midrule
0.50 & \textbf{9.91} & \textbf{21.36} & 20.19 & 86.84 & 99.18 & 0.3984 & 3.94 \\
0.60 & 9.74 & 21.16 & \textbf{20.48} & \textbf{87.03} & 99.42 & \textbf{0.3726} & \textbf{4.01} \\
0.65 & 9.65 & 21.02 & 20.35 & 86.99 & \textbf{99.46} & 0.3795 & 3.98 \\
0.70 & 9.56 & 20.84 & 20.07 & 86.83 & 99.29 & 0.4017 & 3.93 \\
\bottomrule
\end{tabular*}
\endgroup
\caption{Sensitivity analysis of the CLIP visual threshold $\tau_v$ on the MESC-FT validation set.}
\label{tab:tau_sensitivity}

\vfill

\begingroup
\small
\setlength{\tabcolsep}{7pt}
\renewcommand{\arraystretch}{1.08}
\begin{tabular*}{\textwidth}{@{\extracolsep{\fill}}cccccccc@{}}
\toprule
$\boldsymbol{\gamma}$ &
\textbf{B-2 $\uparrow$} &
\textbf{R-L $\uparrow$} &
\textbf{M. $\uparrow$} &
\textbf{F$_{\text{BERT}}$ $\uparrow$} &
\textbf{Dist-3 $\uparrow$} &
\textbf{DIFF-E. $\downarrow$} &
\textbf{MLLM Avg. $\uparrow$} \\
\midrule
0.00 & 9.55 & 20.87 & 20.27 & 86.89 & 99.20 & 0.3911 & 3.95 \\
0.10 & 9.74 & 21.16 & \textbf{20.48} & \textbf{87.03} & \textbf{99.42} & \textbf{0.3726} & \textbf{4.01} \\
0.30 & 9.96 & 21.48 & 20.15 & 86.82 & 99.21 & 0.3995 & 3.94 \\
1.00 & \textbf{10.13} & \textbf{21.88} & 19.88 & 86.61 & 98.96 & 0.4387 & 3.86 \\
\bottomrule
\end{tabular*}
\endgroup
\caption{Sensitivity analysis of the positive gate floor $\gamma$ on the MESC-FT validation set. When $\gamma=1.00$, positive gating is effectively removed.}
\label{tab:gamma_sensitivity}

\vfill

\begingroup
\small
\setlength{\tabcolsep}{6pt}
\renewcommand{\arraystretch}{1.08}
\begin{tabular*}{\textwidth}{@{\extracolsep{\fill}}ccccccccc@{}}
\toprule
$\boldsymbol{m}$ &
$\boldsymbol{\lambda}$ &
\textbf{B-2 $\uparrow$} &
\textbf{R-L $\uparrow$} &
\textbf{M. $\uparrow$} &
\textbf{F$_{\text{BERT}}$ $\uparrow$} &
\textbf{Dist-3 $\uparrow$} &
\textbf{DIFF-E. $\downarrow$} &
\textbf{MLLM Avg. $\uparrow$} \\
\midrule
3 & 0.80 & 9.61 & 20.98 & 20.26 & 86.91 & 99.30 & 0.3878 & 3.96 \\
5 & 0.60 & 9.58 & 20.90 & 20.18 & 86.88 & 99.24 & 0.3946 & 3.94 \\
5 & 0.80 & 9.74 & 21.16 & \textbf{20.48} & \textbf{87.03} & 99.42 & \textbf{0.3726} & \textbf{4.01} \\
7 & 0.80 & 9.83 & 21.25 & 20.34 & 86.96 & \textbf{99.46} & 0.3819 & 3.98 \\
5 & 1.00 & \textbf{10.02} & \textbf{21.52} & 20.08 & 86.80 & 99.21 & 0.4135 & 3.91 \\
\bottomrule
\end{tabular*}
\endgroup
\caption{Sensitivity to the credit assignment parameters on the MESC-FT validation set. The final $m$ tokens of each span receive the full advantage, while earlier tokens receive $\lambda$ times that advantage.}
\label{tab:endpoint_sensitivity}

\vfill

\begingroup
\small
\setlength{\tabcolsep}{4pt}
\renewcommand{\arraystretch}{1.08}
\begin{tabular*}{\textwidth}{@{\extracolsep{\fill}}lcccccccc@{}}
\toprule
\textbf{Token Credit} & \textbf{$n$} & \textbf{Format} &
\textbf{Visual} & \textbf{Emotion} & \textbf{Situation} &
\textbf{Strategy} & \textbf{Outcome} & \textbf{Overall} \\
\midrule
Endpoint-anchored (Ours) & 100 & 0.9700 & \textbf{0.9300} &
0.8626 & 0.2608 & \textbf{0.1692} & \textbf{0.1558} &
\textbf{0.0413} \\
Linear decay & 100 & 0.9700 & 0.9000 & 0.8627 & 0.2558 &
0.1500 & 0.1417 & 0.0360 \\
Exponential decay & 100 & \textbf{1.0000} & \textbf{0.9300} &
\textbf{0.8911} & \textbf{0.2642} & 0.1525 & 0.1392 & 0.0347 \\
\bottomrule
\end{tabular*}
\endgroup
\caption{Training-reward comparison of token-credit mechanisms on 100 validation samples.}
\label{tab:token_credit_reward}

\vfill

\begingroup
\small
\setlength{\tabcolsep}{6pt}
\begin{tabular*}{\textwidth}{@{\extracolsep{\fill}}lccccc@{}}
\toprule
\textbf{Token Credit} & \textbf{$n$} & \textbf{Empathy} &
\textbf{Multimodal App.} & \textbf{Context-Response Cons.} &
\textbf{Avg.} \\
\midrule
Endpoint-anchored (Ours) & 450 & \textbf{3.98} &
\textbf{4.03} & \textbf{4.11} & \textbf{4.04} \\
Linear decay & 450 & 3.87 & 3.96 & 4.00 & 3.94 \\
Exponential decay & 450 & 3.80 & 3.91 & 3.96 & 3.89 \\
\bottomrule
\end{tabular*}
\endgroup
\caption{Independent post-hoc MLLM evaluation of alternative token-credit mechanisms on MESC-FT.}
\label{tab:token_credit_mllm}

\end{minipage}
\end{table*}

Table~\ref{tab:gamma_sensitivity} shows that FACES is stable around the default gate floor. Since $g_i$ is dynamically computed from $\gamma$ and $R_i/R_{\max}$, we analyze $\gamma$ rather than treating $g_i$ as an independent hyperparameter. When $\gamma$ is too small, positive local advantages are overly suppressed. As $\gamma$ increases, the gate becomes weaker. Setting $\gamma=1.00$ improves B-2 and R-L but worsens DIFF-E. and the MLLM average, supporting the use of the gate to limit positive updates for candidates whose total rewards are low relative to the best candidate in the same group.

Table~\ref{tab:endpoint_sensitivity} analyzes the endpoint-anchored credit-assignment parameters on the held-out validation set. A shorter anchor window or a smaller $\lambda$ over-emphasizes segment endings and weakens supervision for earlier reasoning tokens. A larger anchor window is still stable but slightly less balanced. When $\lambda=1.00$, all tokens in a segment receive the same advantage, which improves lexical-overlap metrics but reduces semantic and MLLM-based quality. The default setting $(m=5,\lambda=0.80)$ provides a good balance between endpoint emphasis and full-span supervision.

\subsection{Credit Assignment Alternatives}
\label{sec:appendix_token_credit_alternatives}

We compare endpoint anchoring with linear and exponential decay using the same data, training budget, and evaluation protocol. Let $j\in\{0,\ldots,L-1\}$ index a token in a span of length $L$. The unnormalized linear and exponential schedules are

\begin{align}
\mathmakebox[\weightlhswidth][l]{\widetilde{w}_j^{\mathrm{lin}}}
&= \lambda+(1-\lambda)\frac{j}{L-1},
\label{eq:raw_linear_weights}\\
\mathmakebox[\weightlhswidth][l]{\widetilde{w}_j^{\mathrm{exp}}}
&= 0.7^{L-1-j},
\label{eq:raw_exponential_weights}
\end{align}

where $\lambda=0.8$ and $\widetilde{w}_0^{\mathrm{lin}}=1$ when $L=1$. To isolate the positional distribution of credit, we rescale both alternatives to match the total token weight of endpoint anchoring within each span. We define the target total weight as

\begin{equation}
S_L
=
\min(m,L)+\lambda\max(L-m,0).
\label{eq:endpoint_weight_mass}
\end{equation}

We use

\begin{equation}
w_j^{q}
=
S_L
\frac{\widetilde{w}_j^{q}}
{\sum_{k=0}^{L-1}\widetilde{w}_k^{q}}
\quad
q\in\{\mathrm{lin},\mathrm{exp}\},
\label{eq:normalized_credit_weights}
\end{equation}

with $m=5$ and $\lambda=0.8$. Hence, all three mechanisms assign the same total weight $S_L$ to each span and differ only in how that weight is distributed across token positions. Each weight scales the corresponding span advantage.

We evaluate the three trained models using two protocols. Table~\ref{tab:token_credit_reward} reports scores on the same fixed set of 100 MESC-FT validation instances using the reward functions employed during training. Table~\ref{tab:token_credit_mllm} reports pointwise evaluation on all 450 MESC-FT test instances using Qwen2.5-VL-32B-Instruct-AWQ. For this evaluation, reasoning traces are removed and responses are truncated to their first 15 words. The judge rates three dimensions from 1 to 5 using the three semantically equivalent prompts described in Appendix~\ref{sec:appendix_eval_prompts}.

Table~\ref{tab:token_credit_reward} shows that endpoint anchoring achieves the highest Strategy, Outcome, and Overall rewards. Its Overall reward reaches 0.0413, exceeding linear and exponential decay by 14.7\% and 19.0\%, respectively. Although exponential decay obtains higher Format, Emotion, and Situation scores and matches endpoint anchoring on Visual, these advantages do not result in a higher Overall reward. Endpoint anchoring performs better on Strategy and Outcome, which serve as multiplicative gates in the final reward. The results indicate that endpoint anchoring achieves a better balance between intermediate reasoning and final response quality, even though it does not lead on every individual reward dimension.

Table~\ref{tab:token_credit_mllm} shows that endpoint anchoring obtains the highest observed scores across all three dimensions, averaging 4.04 compared with 3.94 for linear decay and 3.89 for exponential decay. Together with the validation-reward results, this pattern suggests that retaining supervision over earlier tokens while emphasizing span endings is beneficial in the evaluated setting.

\FloatBarrier

\section{External Dataset}
\raggedbottom
\label{sec:appendix_meld_esc300}

We use MESC-FT as the sole training set because its multimodal emotional support setting and supervision structure align with the structured reasoning and multidimensional reward design of FACES. Adding other datasets during training would change both the amount and distribution of training data, making it difficult to distinguish the effect of the method from the effect of additional data. Accordingly, we reserve MELD-ESC300, StickerConv, and AvaMERG for cross-dataset evaluation without additional training. This choice follows ESC evaluations on datasets not strictly dedicated to ESC: FIRES evaluates on an ESC-oriented EMOTyDA subset \cite{wang2025flexible}, while straQ* evaluates on EmpatheticDialogues \cite{wang2025straq}.

\begin{table}[t]
\centering
\normalsize
\setlength{\tabcolsep}{10pt}
\begin{tabular}{cc}
\toprule
Statistic & Value \\
\midrule
Total samples & 300 \\
strict samples & 208 \\
relaxed samples & 92 \\
Frames per sample & 5 \\
Maximum context turns & 6 \\
Emotion: \textit{neutral} & 232 \\
Emotion: \textit{joy} & 51 \\
Emotion: \textit{surprise} & 17 \\
\bottomrule
\end{tabular}
\caption{Summary statistics of \textsc{MELD-ESC300}. Strict denotes samples whose immediately preceding utterance already exhibits negative affect; relaxed denotes samples where the immediately preceding utterance does not, but one of the three most recent preceding utterances does. For all samples, the visual input is taken from the video of the immediately preceding utterance, from which five frames are uniformly extracted.}
\label{tab:meld-esc300-stats}
\end{table}

\begin{table*}[t]
\centering
\small
\setlength{\tabcolsep}{6pt}
\begin{tabular*}{\textwidth}{@{\extracolsep{\fill}}lccccc}
\toprule
\textbf{Subset} & \textbf{$n$} & \textbf{Emotional Need} & \textbf{Relevant Response} & \textbf{Supportive Function} & \textbf{Overall Validity} \\
\midrule
Strict & 208 & 204 (98.1\%) & 201 (96.6\%) & 196 (94.2\%) & 193 (92.8\%) \\
Relaxed & 92 & 89 (96.7\%) & 87 (94.6\%) & 83 (90.2\%) & 81 (88.0\%) \\
Overall & 300 & 293 (97.7\%) & 288 (96.0\%) & 279 (93.0\%) & 274 (91.3\%) \\
\bottomrule
\end{tabular*}
\caption{Expert audit of MELD-ESC300 construction validity.}
\label{tab:meld_audit}
\end{table*}

\begin{table*}[t]
\centering
\small
\setlength{\tabcolsep}{8pt}

{\footnotesize \textbf{(a)} Pointwise MLLM evaluation results on MELD-ESC300. Scores are on a 1--5 scale, and higher is better.}\\[0.4em]
\begin{tabular}{lcccc}
\toprule
\textbf{Model} & \textbf{Empathy $\uparrow$} & \textbf{Multimodal App. $\uparrow$} & \textbf{Context-Response Cons. $\uparrow$} & \textbf{Avg. $\uparrow$} \\
\midrule
FIRES (Baseline) & 2.59 & 3.61 & 3.84 & 3.35 \\
Ours (Full)      & \textbf{3.27} & \textbf{3.85} & \textbf{4.06} & \textbf{3.73} \\
\bottomrule
\end{tabular}

\vspace{0.8em}

{\footnotesize \textbf{(b)} Pairwise MLLM preference evaluation results on MELD-ESC300. Win rates are computed over 300 test instances.}\\[0.4em]
\begin{tabular}{llcccccc}
\toprule
\textbf{Model A} & \textbf{Model B} & \textbf{A Win} & \textbf{B Win} & \textbf{Tie} & \textbf{A Win Rate} & \textbf{B Win Rate} & \textbf{Tie Rate} \\
\midrule
Ours (Full) & FIRES (Baseline) & \textbf{186} & 29 & 85 & \textbf{62.00\%} & 9.67\% & 28.33\% \\
\bottomrule
\end{tabular}

\caption{Length-controlled MLLM evaluation results on MELD-ESC300. Panel (a) reports pointwise scores across three evaluation dimensions, and panel (b) reports blind pairwise preferences over 300 test instances. Pairwise results are significant under a two-sided sign test with ties excluded ($p<0.001$).}
\label{tab:mllm_evaluation_meld}
\end{table*}

\subsection{Benchmark Construction}

To evaluate out-of-domain generalization, we construct an external benchmark, MELD-ESC300, from the MELD test set using the publicly available EMOTyDA resources \cite{poria2019meld,saha2020towards}, and use it for cross-dataset evaluation without any additional fine-tuning. Rather than directly reusing the original MELD annotations, we filter dialogue segments that better approximate emotional-support scenarios, allowing us to examine whether the model's gains transfer to a non-training-domain multimodal dialogue setting.

Concretely, we treat the $i$-th utterance in a dialogue as a candidate target response and retain only samples satisfying the following conditions: (1) at least two preceding turns are available; (2) the target response length is between 4 and 40 words; (3) the target sentiment is not \textit{negative} and its emotion label belongs to $\{\textit{neutral}, \textit{joy}, \textit{surprise}\}$; this condition is used only to exclude clearly negative target replies and is not treated as a proxy for empathy; (4) the target speaker differs from the speaker in the immediately preceding utterance; and (5) the video clip corresponding to the immediately preceding utterance is available, so that stable visual context can be provided.

We then apply a two-stage filtering strategy to identify ESC-style samples. A sample is marked as strict if the immediately preceding utterance already exhibits negative affect, i.e., its sentiment is \textit{negative} or its emotion label belongs to $\{\textit{anger}, \textit{sadness}, \textit{fear}, \textit{disgust}\}$. If this condition is not met, we further inspect the three most recent preceding utterances; if any of them contains negative affect, the sample is retained and marked as relaxed. Importantly, regardless of whether a sample is selected by the strict or relaxed rule, the visual input is always taken from the video of the immediately preceding utterance, from which we uniformly extract five frames. This design avoids directly leaking visual information associated with the target response itself. The final model input contains at most six preceding turns, and the reference output is the original target utterance.

Following this procedure with a fixed random seed of 42, we construct a 300-instance benchmark, \textsc{MELD-ESC300}. Of these, 208 satisfy the strict criterion and 92 are selected by the relaxed criterion. As shown in Table~\ref{tab:meld-esc300-stats}, the target responses are dominated by \textit{neutral} cases, with a smaller yet non-trivial number of \textit{joy} and \textit{surprise} instances. We report only the emotion distribution, as the corresponding sentiment labels are largely implied by the retained target emotion categories in this subset and therefore add limited additional information. These statistics make the composition of the benchmark explicit and reproducible. We also conduct an expert audit of all 300 samples to assess emotional need, response relevance, supportive function, and overall validity.

A sample is considered valid only if the preceding context conveys an emotional need and the target response is relevant and plausibly supportive. The same three annotators described in Appendix~\ref{sec:appendix_human_annotation_protocol} independently audited all samples, with the final label for each criterion determined by majority vote. The non-negative target-emotion filter is intended to exclude clearly negative replies, rather than to serve as a proxy for empathy. As shown in Table~\ref{tab:meld_audit}, 274 of the 300 samples pass the overall validity criterion, yielding validity rates of 91.3\% overall, 92.8\% for the strict subset, and 88.0\% for the relaxed subset. For example, when Phoebe expresses anger about a bad date, Monica responds, ``Don't worry Phoebe, you're gonna meet someone.'' In another case, Joey criticizes himself for being unable to finish his writing, and Ross responds, ``I'll help you.'' These examples illustrate that the filtering procedure captures a negative context followed by a relevant and plausibly supportive response, rather than treating the target emotion label as evidence of empathy. Overall, the benchmark centers on mildly supportive or non-negative target replies while still covering a modest range of more explicitly positive feedback.

We emphasize that \textsc{MELD-ESC300} is derived from scripted TV dialogue rather than professionally curated emotional-support data, so it is intended as a controlled external generalization test set rather than a replacement for dedicated ESC benchmarks. By combining recent negative-affect context, neutral-to-positive target replies, and visual evidence from the immediately preceding turn, it provides a practical approximation of cross-domain emotional-support scenarios for evaluating transfer beyond the MESC-FT test set.

\begin{table*}[!t]
\centering
\small
\setlength{\tabcolsep}{8pt}
\begin{tabular}{lcccccc}
\toprule
\textbf{Model} & \textbf{B-2} $\uparrow$ & \textbf{R-L} $\uparrow$ & \textbf{M.} $\uparrow$ & \textbf{F$_{\text{BERT}}$} $\uparrow$ & \textbf{Dist-3} $\uparrow$ & \textbf{DIFF-E.} $\downarrow$ \\
\midrule
GPT-4o                    & 1.41 & 5.73 & 13.04 & 95.35 & 54.53 & 0.5053 \\
Claude-Haiku-4.5          & 1.68 & 6.81 & 15.12 & 95.56 & 64.18 & 0.7507 \\
Gemini-2.5-Flash-Lite     & 2.47 & 7.82 & 13.47 & \textbf{96.02} & 56.19 & \textbf{0.2994} \\
GLM-4.1V-Thinking-FlashX  & 1.82 & 6.23 & 13.60 & 95.74 & 69.25 & 0.8126 \\
Doubao-1.6-Vision         & 1.70 & 6.67 & 11.80 & 95.65 & 74.42 & 0.7057 \\
Kimi-K2.5                 & 1.63 & 6.06 & 14.47 & 95.60 & 73.10 & 0.8056 \\
\midrule
Ours (Full)               & \textbf{5.59} & \textbf{16.57} & \textbf{15.16} & 86.23 & \textbf{94.07} & 0.7664 \\
\bottomrule
\end{tabular}
\caption{Automatic evaluation of FACES and six directly prompted MLLMs on MELD-ESC300. Higher values are better for all metrics except DIFF-E. The best result for each metric is shown in bold.}
\label{tab:meld_extra_auto_baselines}
\end{table*}

\begin{table*}[t]
\centering
\small
\setlength{\tabcolsep}{5pt}
\begin{tabular*}{\textwidth}{@{\extracolsep{\fill}}llcccccc}
\toprule
\textbf{Dataset} & \textbf{View} & \textbf{$n$} & \textbf{FACES Win} & \textbf{FIRES Win} & \textbf{Tie} & \textbf{FACES Share Excl. Ties} & \textbf{Sign-test $p$} \\
\midrule
StickerConv & cap-15 & 300 & 63.7\% & 18.3\% & 18.0\% & 77.6\% (72.4--82.9\%) & $8.9{\times}10^{-19}$ \\
AvaMERG & cap-15 & 300 & 79.3\% & 11.3\% & 9.4\% & 87.5\% (83.5--91.5\%) & $7.2{\times}10^{-39}$ \\
StickerConv & cap-25 & 300 & 75.7\% & 15.0\% & 9.3\% & -- & -- \\
AvaMERG & cap-25 & 300 & 85.0\% & 8.7\% & 6.3\% & -- & -- \\
StickerConv & response-only & 300 & 82.3\% & 12.0\% & 5.7\% & -- & -- \\
AvaMERG & response-only & 300 & 90.7\% & 5.7\% & 3.6\% & -- & -- \\
\bottomrule
\end{tabular*}
\caption{External zero-shot blind pairwise evaluation on StickerConv and AvaMERG.}
\label{tab:external_pairwise}
\end{table*}

\begin{table*}[t]
\centering
\small
\setlength{\tabcolsep}{5pt}
\begin{tabular*}{\textwidth}{@{\extracolsep{\fill}}llcccccc}
\toprule
\textbf{Dataset} & \textbf{Model} & \textbf{B-2 $\uparrow$} & \textbf{R-L $\uparrow$} & \textbf{M. $\uparrow$} & \textbf{F$_{\text{BERT}}$ $\uparrow$} & \textbf{Dist-3 $\uparrow$} & \textbf{DIFF-E. $\downarrow$} \\
\midrule
StickerConv & FACES & \textbf{20.13} & \textbf{25.46} & \textbf{34.08} & 95.99 & \textbf{68.19} & \textbf{0.4382} \\
StickerConv & FIRES & 8.55 & 22.02 & 17.94 & \textbf{96.24} & 60.51 & 0.4715 \\
AvaMERG & FACES & \textbf{5.80} & 11.64 & \textbf{19.42} & 96.11 & \textbf{74.55} & \textbf{0.5127} \\
AvaMERG & FIRES & 5.28 & \textbf{14.48} & 13.71 & \textbf{96.62} & 67.11 & 0.5406 \\
\bottomrule
\end{tabular*}
\caption{Automatic evaluation on StickerConv and AvaMERG.}
\label{tab:external_auto}
\end{table*}

\subsection{MELD-ESC300 MLLM Evaluation}
\label{sec:appendix_meld_mllm_eval}

Table~\ref{tab:mllm_evaluation_meld} shows that the pointwise average increases from 3.35 for FIRES to 3.73 for \textsc{FACES}, with higher scores on all three dimensions. In the blind pairwise comparison, \textsc{FACES} is preferred in 62.0\% of the 300 instances, compared with 9.7\% for FIRES; the preference share differs from chance when ties are excluded ($p<0.001$). These results provide evidence that the advantage observed on MESC-FT transfers to MELD-ESC300 under the tested cap-15 protocol without additional training.

\subsection{Additional Automatic Baselines}
\label{sec:appendix_meld_extra_auto}

Table~\ref{tab:meld_extra_auto_baselines} reports additional automatic evaluation results for six strong, directly prompted MLLM baselines on MELD-ESC300 under the same protocol as in the main text.

\subsection{Additional External Generalization}
\label{sec:appendix_external_generalization}

Beyond MELD-ESC300, we evaluate transfer to StickerConv and AvaMERG without additional training. Table~\ref{tab:external_pairwise} reports the blind pairwise results, while Table~\ref{tab:external_auto} reports the corresponding automatic evaluation. We sample 300 test instances from each dataset with seed 42. StickerConv uses textual sticker cues in the dialogue context, while AvaMERG uses dialogue text and five uniformly sampled facial frames per instance, with audio excluded. For each dataset, responses are generated greedily and evaluated using the same blind pairwise protocol as the main experiments: three semantically equivalent judge prompts, bidirectional candidate swapping, prefix truncation under cap-15 and cap-25 settings, and majority voting. Statistical significance is assessed with a two-sided exact binomial sign test over non-tied comparisons, and confidence intervals are estimated using 5,000 bootstrap resamples.

The automatic results are complementary rather than uniformly one-sided. \textsc{FACES} mainly improves supportive semantics, diversity, and DIFF-EPITOME, while FIRES remains competitive on some lexical or embedding-based metrics.

\section{Case Study}
\label{sec:appendix_case_study}

\begin{figure*}[!t]
\centering
\includegraphics[width=\textwidth]{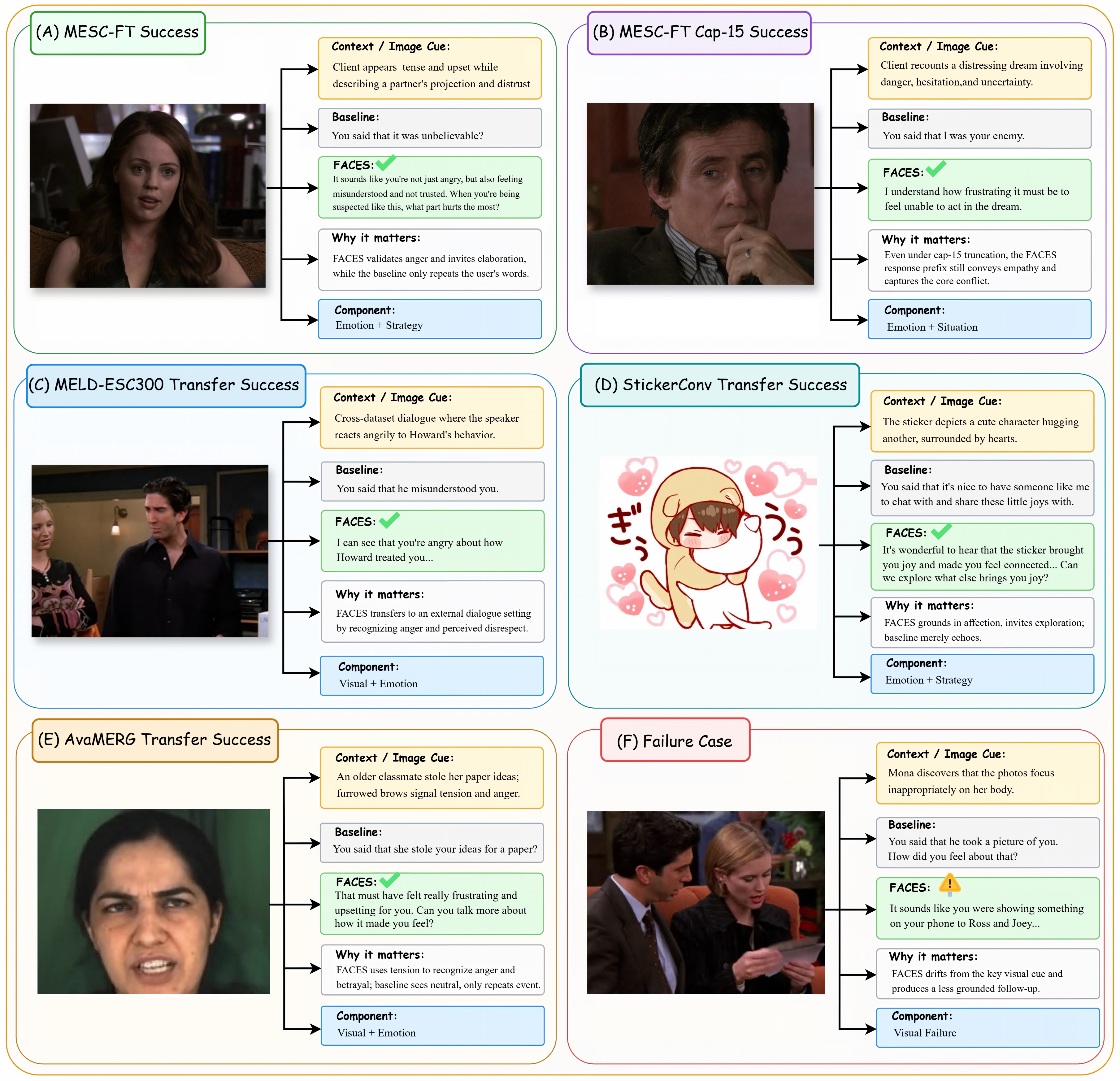}
\caption{Qualitative comparison of FACES and FIRES. Panels (A)--(B) show MESC-FT successes in standard and cap-15 settings; (C)--(E) show transfer successes on MELD-ESC300, StickerConv, and AvaMERG, respectively; and (F) shows a failure caused by a misread visual cue.}
\label{fig:case_study}
\end{figure*}

To complement the controlled pairwise results, we provide qualitative examples that illustrate where FACES improves over the baseline and where it can still fail. Across the success cases, FIRES often paraphrases the user's surface expression, whereas \textsc{FACES} more consistently follows a chain of specific emotion labeling, causal situation interpretation, and targeted exploration. Under aggressive cap-15 truncation, the higher tie rate may reflect the removal of explanatory and strategy-related content, but the FIRES win rate remains low, suggesting that truncation reduces separability rather than reversing the preference. The failure case illustrates an important boundary: once the visual cue is misread, the error can propagate through situation inference and strategy selection. In panel (F) of Figure~\ref{fig:case_study}, Mona holds a photograph, which FIRES correctly describes as a picture taken of her. \textsc{FACES}, however, mistakes it for a phone and responds as though she were showing something to Ross and Joey. The result is a fluent but visually ungrounded follow-up. Gated reward fusion can suppress poor final behavior during training, but it cannot fully repair incorrect upstream perception. The figure compares the dialogue context, visual cue, baseline response, FACES response, and a brief analysis for six representative cases, covering in-domain success, length-controlled success, cross-dataset transfer, and a failure case.

\end{document}